\documentclass[10pt,twocolumn,letterpaper]{article}

\usepackage[algorithms]{wacv}
\definecolor{wacvblue}{rgb}{0.21,0.49,0.74}
\usepackage[pagebackref,breaklinks,colorlinks,allcolors=wacvblue]{hyperref}

\usepackage{pifont} 
\usepackage{colortbl}
\usepackage{graphicx}
\usepackage{booktabs}
\usepackage{enumitem}
\usepackage{bm}
\usepackage{booktabs}
\usepackage{multirow}
\usepackage{makecell}
\usepackage{subcaption}
\usepackage[percent]{overpic}
\usepackage{fvextra}

\usepackage[accsupp]{axessibility}
\newcommand{\method}{\textsc{InterPartAbility}}

\title{
\begin{minipage}[c]{0.08\textwidth}
\centering
\includegraphics[height=1.9cm]{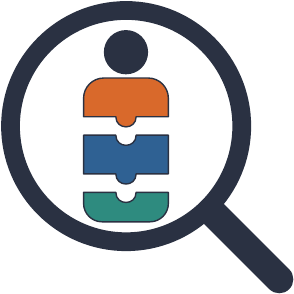}
\end{minipage}
\hfill
\begin{minipage}[c]{0.913\textwidth}
\centering
\method: Phrase-Region Grounding for Interpretable Text-to-Image Person Re-Identification
\end{minipage}
}

\author{Shakeeb Murtaza\textsuperscript{1,}\thanks{Equal contribution} , Aryan Shukla\textsuperscript{1,}\footnotemark[1] , Rajarshi Bhattacharya\textsuperscript{1}, Maguelonne Heritier\textsuperscript{2}, Eric Granger\textsuperscript{1}\\
\textsuperscript{1}LIVIA, Dept. of Systems Engineering, ETS Montreal, Canada \\
\textsuperscript{2}Genetec Inc.\\
}

\definecolor{cred}{HTML}{FF0000}
\definecolor{cblue}{HTML}{4F81FF}
\definecolor{cgreen}{HTML}{029500}

\AtBeginDocument{%
  \setlength{\abovedisplayskip}{2pt plus 1pt minus 1pt}%
  \setlength{\belowdisplayskip}{2pt plus 1pt minus 1pt}%
  \setlength{\abovedisplayshortskip}{1pt plus 1pt}%
  \setlength{\belowdisplayshortskip}{2pt plus 1pt minus 1pt}%
}
\makeatletter
\def\wacvsection{\@startsection {section}{1}{\z@}
   {-8pt plus -2pt minus -2pt}{5pt} {\large\bf}}
\def\wacvsubsection{\@startsection {subsection}{2}{\z@}
   {-6pt plus -1pt minus -1pt}{3pt} {\elvbf}}
\makeatother

\begin{document}
\maketitle

\begin{abstract}
Text-to-image person re-identification (TI-ReID) relies on natural-language text descriptions to retrieve top matching individuals from a gallery of reference images. While recent large vision-language models (VLMs) achieve strong retrieval performance, their decisions remain largely uninterpretable. Existing interpretability approaches in TI-ReID rely solely on slot-attention to highlight attended regions, but fail to reliably bind visual regions to semantically meaningful concepts, limiting interpretation to qualitative visualizations over a restricted vocabulary. This paper introduces \method, an interpretable TI-ReID method that performs explicit part-wise matching and enables phrase-region grounding. Unlike parameter-heavy slot-attention methods that yield only qualitative interpretability, our open-vocabulary patch-phrase interaction module (PPIM) guides a standard TI-ReID model with concept-level phrases. Concept-based part phrases provide evidence that encourages the model to attend to the corresponding local image regions. \method{} further leverages CLIP ViT self-attention to produce spatially concentrated patch activations aligned with each part-level phrase, yielding grounded explanation maps. Finally, a quantitative interpretability protocol for TI-ReID is introduced that extends current perturbation-based evaluation metrics into the TI-Reid domain. This includes a counterfactual region removal that measures retrieval degradation when top-ranked explanatory regions are removed. Empirical results on three challenging benchmarks show that \method{} can achieve SOTA interpretability performance under these metrics, while sustaining competitive retrieval accuracy. 
\end{abstract}

\begin{figure}[!t]
    \centering
    \includegraphics[width=0.93\linewidth]{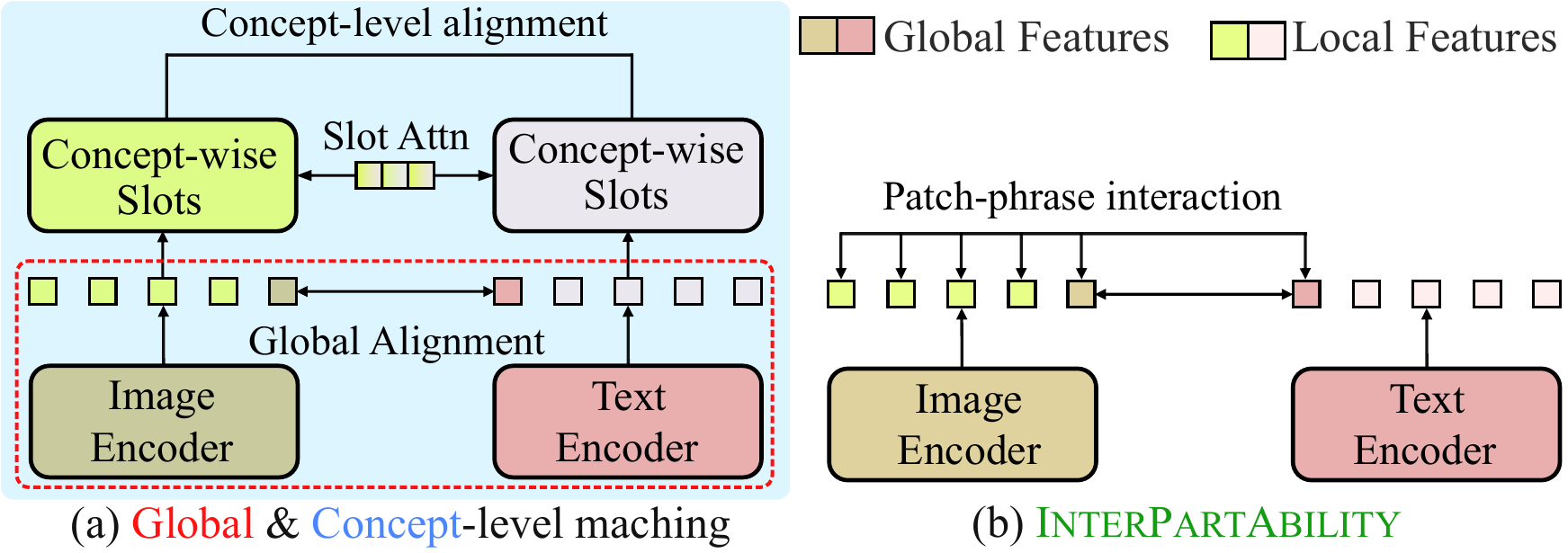}
    \caption{
    TI-ReID alignment paradigms. (a) \textcolor{cred}{\textit{Global matching}}: VLM-based methods (e.g., CLIP) produce global image-text similarity, offering no insight into which regions. \textcolor{cblue}{\textit{Concept-level matching}} (i.e., PLOT~\cite{park2024plot} and DiCo~\cite{kim2026dico}): slot attention decomposes features into concept regions but fails to bind slots to specific textual phrases, yielding unlabeled qualitative visualizations with high computational cost due to slots. (b) \textcolor{cgreen}{\textit{\method{} (ours)}}: patch embeddings interact directly with open-vocabulary phrase embeddings via a lightweight patch-phrase interaction module, explicitly grounding each appearance phrase to its corresponding image region without architectural overhead.
    }
    \label{fig:teaser}
\end{figure}

\section{Introduction}

TI-ReID aims to retrieve a ranked set of reference images from a gallery that best match a natural-language description of a person (e.g., ``a man in a red shirt carrying a black backpack''). The gallery may contain images captured across different camera viewpoints, with variations in pose, illumination, resolution, blur, etc. TI-ReID operates on two data modalities, namely textual and visual, presenting a unique set of challenges. The model must learn to form associations between semantic concepts in textual space and their corresponding visual evidence, requiring reasoning over fine-grained local attributes and their spatial grounding. Another challenge is that real-world TI-ReID annotations exhibit noisy correspondence, where the ambiguities lead to partial or total misalignment between modalities~\cite{qin2024noisy,zuo2024ufinebench}. Part-level grounding is therefore essential.

VLMs such as CLIP have dramatically improved text-to-image retrieval~\cite{radford2021learning}. By training a dual-encoder architecture on millions of image-text pairs, CLIP learns a shared embedding space that places semantically similar images and texts closer together, facilitating cross-modal alignment. Several recent works have adapted CLIP to person TI-ReID by fine-tuning it on pedestrian datasets such as CUHK-PEDES ~\cite{li2017person}, ICFG-PEDES~\cite{ding2021semantically}, and RSTPReID ~\cite{rstpreid}. However, most state-of-the-art approaches remain \emph{global} and largely \emph{uninterpretable} -- they aggregate the entire image into a single embedding and the entire caption into another global embedding, allowing only holistic similarity matching. This global matching process does not explain why a particular image was retrieved. This limitation is critical for TI-ReID. Given a match, it is important to understand whether the retrieval was driven by the red shirt, the black backpack, or some spurious background correlation. If the model fails, we should know whether an attribute was mislocalized, a relevant phrase was ignored, or it relied on an incidental cue. Global image-text similarity does not provide such insight.

Although part-based and interpretable methods have been extensively explored in image-to-image ReID,~\cite{alehdaghi2025beyond,heritier2025exam}, interpretability in TI-ReID remains underdeveloped. Recent methods like PLOT~\cite{park2024plot} and DiCo~\cite{kim2026dico} inherit slot-attention mechanisms~\cite{locatello2020object} to produce attention visualizations. However, these explanations remain largely qualitative and rely on a fixed number of slots to represent activated regions. Using a predefined number of slots introduces structural limitations: the capacity to interpret depends on the number of slots, potentially merging multiple semantic regions into one. Increasing the number of slots may improve granularity, but leads to parameter growth that increases computational cost. Beyond architectural limitations, state-of-the-art TI-ReID interpretability methods are evaluated almost exclusively through qualitative visualizations. While attention maps can appear intuitive, they do not provide a measurable notion of whether the highlighted regions are causally responsible for the retrieval decision. No quantitative mechanism exists to verify that an explanation faithfully reflects the model’s reasoning or merely correlates with it.

To address these challenges, this paper introduces \method, a CLIP-based framework that leverages internal attention mechanisms of TI-ReID models to produce concept-level, part-aware visualizations without relying on slot-based decomposition. Fig.\ref{fig:teaser} illustrates this evolution from global to slot-based, and to our phrase-conditioned alignment strategy. Global paradigms collapse all part-based regional concepts into a single vector, thereby stopping the model from learning local patch-phrase alignment. With slot-based paradigms, there is no supervisory signal that forces models to learn localized, region-specific information for particular concept parts. Instead, it offloads this task onto the inherent disentanglement capabilities of slot-attention, at the expense of a large increase in parameters. To address this problem, our patch-phrase interaction module (PPIM) exploits part-phrases to learn local representations by contrasting phrase-conditioned region representations across identities. It enables the visual encoder to ground each appearance phrase to its corresponding image region, yielding part-level interpretation maps at no additional trainable parameters. Moreover, to assess whether the regions highlighted by our method truly interpret the retrieval decisions of our model, we introduce a counterfactual region-removal evaluation protocol.

\noindent \textbf{The key contributions of our work are:}
\textbf{(1)} \method, a CLIP-based TI-ReID method that encourages explicit patch–phrase alignment, enabling concept-level and part-aware grounding without relying on slot-based decomposition. 
\textbf{(2)} A lightweight PPIM that leverages internal attention to associate textual phrases with localized visual regions, encouraging interpretable part-level matching during training. 
\textbf{(3)} A counterfactual region-removal evaluation protocol that quantitatively measures the causal influence of localized regions on retrieval performance, providing a principled way to assess explanation faithfulness in TI-ReID models. 

\section{Related Work}
\noindent \textbf{Text-to-Image Person Re-ID. } TI-ReID aims to retrieve images from a gallery given a free-form natural language description. State-of-the-art methods can be broadly grouped by their alignment granularity and backbone architecture. Early approaches \cite{li2017person, zhang2018deep, gao2021contextual} employed modality-specific encoders (e.g., CNN-RNN) with global matching objectives. Subsequent work introduced part-level alignment through external supervision. For instance, PMA \cite{jing2020pose} utilized pose keypoints for defining different body regions. Similarly, ViTAA \cite{wang2020vitaa} employed a human parsing network to disentangle attribute-level features. SSAN \cite{ding2021semantically} focused on implicit part correspondences via a non-local network without explicit pose or parsing supervision. With the advent of VLM pre-training, CLIP-based methods \cite{radford2021learning} have become dominant. IRRA \cite{jiang2023cross} proposed a CLIP-based framework along with masked language modelling to fuse visual cues into textual tokens, paired with a similarity distribution matching loss. CFine \cite{yan2023clip} and TBPS-CLIP \cite{cao2024empirical} introduced coarse-to-fine and systematic CLIP adaptation strategies, respectively. Recently, MUM \cite{zhao2024unifying} proposed modelling features as Gaussian distributions to capture matching uncertainty. Similarly, RDE \cite{qin2024noisy} and CFAM \cite{zuo2024ufinebench} proposed utilizing noisy correspondences and granularity by selecting tokens at the token level and using longer descriptions, respectively. Interactive paradigms have also emerged, with ICL \cite{qin2025human} and ChatPR \cite{bai2025chat} utilizing multimodal LLMs for dialogue-based query refinement. 

\noindent \textbf{Interpretable Person Re-Identification. } 
QAConv \cite{liao2020interpretable} introduced query-adaptive convolution kernels to highlight local correspondences between query and gallery images for visualizing part-level matching. Recently, slot attention-based methods \cite{park2024plot, kim2026dico} proposed an interpretability paradigm for TIReID. For instance, \cite{park2024plot} employs a part-discovery module based on slot attention \cite{locatello2020object} to discover part-regions. PLOT \cite{kim2026dico} introduces a hierarchical slot-concept architecture to reduce the complexity and group related body parts into different slots. While \cite{park2024plot} and \cite{kim2026dico} can highlight discovered parts, they are unable to bind slots to semantic textual concepts. This limits explanations to example-based comparisons rather than textual grounding. Concept-based interpretability methods have been proposed for mining semantically meaningful concepts from learned representations \cite{kim2018interpretability, chen2019looks, nauta2023pip}. For instance, \cite{kim2018interpretability} introduces Concept Activation Vectors (CAVs) for global concept-level explanations. Similarly, Prototype-based methods such as ProtoPNet \cite{chen2019looks} and PIPNet \cite{nauta2023pip} provide example-based reasoning through prototypical image patches, but they struggle with semantic consistency across instances. To deal with this issue, ExaM \cite{heritier2025exam} proposed an interpretable-by-design model to discover CAV using object-centric representation learning. Similarly, PCMNet \cite{alehdaghi2025beyond} combines part discovery with concept mining. Firstly, it learns dynamic part-level prototypes using cluster center loss, then utilizes DBSCAN clustering within each class to extract concept prototypes. These methods can produce a reliable explanation, yet both are proposed for image classification and single-modality ReID. They do not address cross-modal text-image matching or provide phrase-level grounding of visual regions.

A critical issue with state-of-the-art methods for interpretable person-ReID is that they only rely on qualitative evaluation, e.g., attention maps for PLOT \cite{park2024plot} and DiCo \cite{kim2026dico}, concept-activation maps for ExaM\cite{heritier2025exam}, and prototype visualizations for PCMNet\cite{alehdaghi2025beyond}. In addition, these methods are unable to produce explicit phrase-region grounding or part-level faithfulness of explanations in a counterfactual setting for the retrieval task. \method addresses this gap in two ways: (i) it produces explicit part-level phrase-region correspondences that ground each textual phrase to specific image regions, and (ii) it adapts interpretability metrics, including counterfactual visual explanation via top-ranked region masking, to quantitatively evaluate the reliability of explanations.

% ============================================================
%  Section: Proposed Method
% ============================================================
\section{Proposed \method}
\label{sec:method}

\begin{figure*}[t]
    \centering
    \includegraphics[width=0.9\linewidth,trim={25pt 7pt 20pt 6pt},clip]{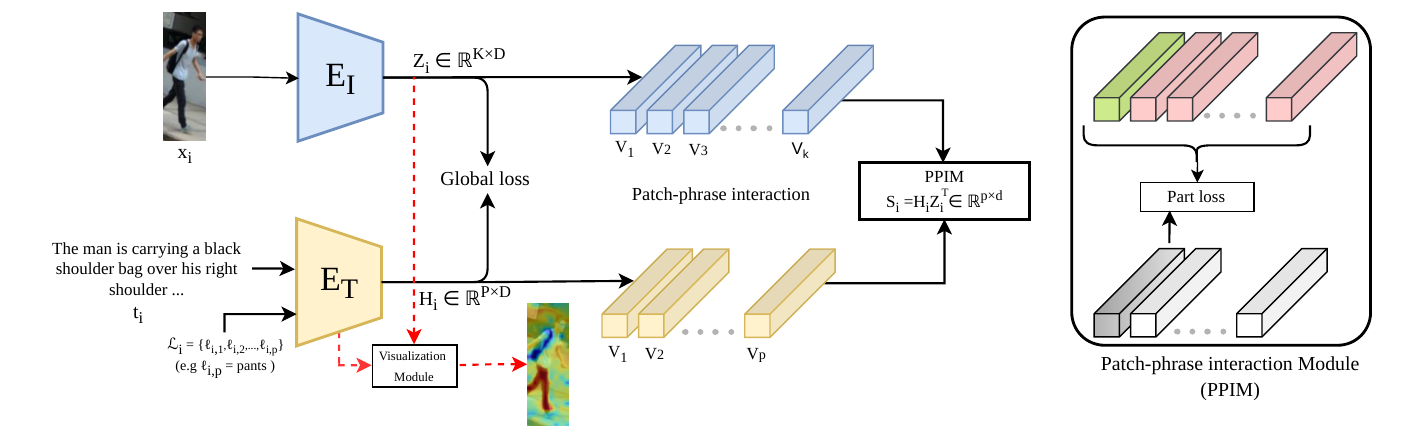}
    \caption{Overview of \method. An image and caption are encoded by CLIP encoders $E_I$ and $E_T$. Global embeddings are trained with the base retrieval objective $\mathcal{L}_{\text{base}}$. The image encoder additionally produces patch embeddings $\bm{Z}_i \in \mathbb{R}^{K \times D}$. Each appearance phrase $\ell_{i,p}$ is encoded into a phrase embedding $\bm{H}_i \in \mathbb{R}^{P \times D}$. In the figure, the tokens labelled $V_1,\dots,V_k$ in the visual stream are the per-patch embeddings $\{\bm{z}_{i,k}\}_{k=1}^{K}$ (rows of $\bm{Z}_i$), and those labelled $V_1,\dots,V_p$ in the text stream are the per-phrase embeddings $\{\bm{h}_{i,p}\}_{p=1}^{P}$ (rows of $\bm{H}_i$). The Patch-Phrase Interaction Module computes phrase-patch similarity and softly aggregates patch features into phrase-conditioned region representations. These are trained with a cross-image contrastive objective ($\mathcal{L}_{\text{part}}$) together with a coverage regulariser. At inference, phrase-patch similarity scores yield spatial heatmaps for each phrase.
    }
    \label{fig:mainarch}
\end{figure*}

Our objective is to develop a person TI-ReID model based on CLIP that (1) achieves competitive retrieval performance, (2) provides part-wise explanations for retrieval outcomes, and (3) enables quantitative evaluation of interpretability. To this end, a part-based training objective is introduced that encourages phrase-conditioned phrase-region alignment within a CLIP-based TI-ReID model, while leaving the global retrieval objective unchanged. In addition, a counterfactual evaluation protocol is proposed to quantify the faithfulness of the resulting interpretations.
Fig.\ref{fig:mainarch} shows our architecture. The image and its corresponding query, and phrases are fed to the image \(E_I\) and the text \(E_T\) encoders respectively. Standard TI-ReID global contrastive losses are calculated, we denote them as \(L_{base}\) then we take the phrase features \(v_i\)'s and patch features \(V_j\) and compute the patch-phrase simialrity matrix between them, this matrix is then used by our part loss \(L_{part}\) to supervise the entire network. To generate the interpretability maps, we use the phrase and patch features and then calculate how strongly each phrase attends to the different regions in the image. After part-based supervision in our method, this local patch-phrase interaction is improved, as shown in Fig.\ref{fig:qual_heatmaps}.

\noindent\textbf{Notation.}
Let $\mathcal{D}=\{(x_i, t_i, y_i, \mathcal{L}_i)\}_{i=1}^{N}$ denote the training set, where $x_i$ is an image, $t_i$ is a caption, $y_i$ is the person identity, and $\mathcal{L}_i = \{\ell_{i,1}, \dots, \ell_{i,P}\}$ is a set of short open-vocabulary appearance phrases (e.g., ``red hoodie'', ``black backpack''). Here, $P$ denotes the maximum number of phrases per sample after padding. Phrase positions do not correspond to predefined semantic attributes; each sample contains its own unordered set of phrases, and unused positions are masked during training. In our setting, phrases are automatically generated using Qwen3 MLLM~\cite{yang2025qwen3}.

A CLIP-style dual encoder is employed, consisting of an image encoder $E_I$ and a text encoder $E_T$. For each image-caption pair, the model produces:
(i) a global image embedding $\bm{v}_i \in \mathbb{R}^D$,
(ii) a global text embedding $\bm{u}_i \in \mathbb{R}^D$, and
(iii) a sequence of patch embeddings $\{\bm{z}_{i,k}\}_{k=1}^{K} \subset \mathbb{R}^D$ from the final visual transformer layer, excluding the \texttt{[CLS]} token.
Each phrase $\ell_{i,p}$ is encoded independently by $E_T$ to obtain a phrase embedding $\bm{h}_{i,p} \in \mathbb{R}^D$ (taking the \texttt{[EOS]} token). All embeddings are $\ ell_2$-normalized.

% ----------------------------
\noindent \textbf{Base Retrieval Objectives}.
% ----------------------------
%
The base model follows the RDE and ICL frameworks~\cite{qin2024noisy, qin2025human}, which optimize global image-text retrieval using Triplet Alignment Loss (TAL) on basic global embeddings (BGE) and token selection embeddings (TSE). We retain this global objective unchanged and denote it as $\mathcal{L}_{\text{base}}$. The proposed part-based loss operates exclusively on patch embeddings and is added on top of $\mathcal{L}_{\text{base}}$.

% ----------------------------
%  Figure: Method Overview (use your existing diagram)
% ----------------------------

% ----------------------------
\subsection{Patch-Phrase Interaction Module (PPIM)}
\label{sec:part_loss}
% ----------------------------
Global retrieval models match a whole caption to a whole image. The caption \texttt{"person wearing a red hoodie, black pants, and brown shoes carrying a backpack"} describes multiple distinct spatial regions. A global embedding collapses all of this into one vector, which means the model cannot know which part of its representation is responsible for the match. This is both an interpretability and a representational problem. The model has no inductive pressure to disentangle person attributes spatially. The proposed part loss introduces exactly this pressure. By constructing phrase-conditioned region representations and contrasting them across images, the part loss introduces patch-level supervision that encourages the visual encoder to produce spatial features that are discriminative with respect to specific appearance attributes, rather than relying solely on global identity-level representations. 
Fig.\ref{fig:mainarch} presents an overview of \method, where the part loss decomposes into three components: (i) phrase-to-patch similarity, (ii) soft phrase assignment with phrase-region aggregation, and (iii) cross-image contrastive supervision with coverage regularization.

\noindent \textbf{Phrase-to-Patch Similarity.}
For each image $i$, let $\bm{Z}_i \in \mathbb{R}^{K \times D}$ denote patch embeddings and $\bm{H}_i \in \mathbb{R}^{P \times D}$ phrase embeddings. We compute the phrase-to-patch similarity matrix, where $\bm{S}_i[p,k] = \langle \bm{h}_{i,p}, \bm{z}_{i,k} \rangle$ measures cosine similarity between phrase $p$ and patch $k$ within the same image of sample $i$:
$\bm{S}_i = \bm{H}_i \bm{Z}_i^\top \in \mathbb{R}^{P \times K}$

\noindent \textbf{Soft Phrase Assignment and Region Aggregation.}
For each patch $k$, we apply a temperature-scaled softmax over phrases:
\begin{equation}
a_{i,k,p} =
\frac{\exp(\bm{S}_i[p,k] / \tau_{\text{part}})}
{\sum_{p'} \exp(\bm{S}_i[p',k] / \tau_{\text{part}})}
\label{eq:patch_assign}
\end{equation}
where $\tau_{\text{part}}$ controls assignment sharpness. The softmax is over phrases rather than patches: every patch is softly partitioned among phrases. Each patch may contain more than one attribute. For instance, a patch may contain the \textit{shoes} as well as the \textit{pants} of the person, so this patch can be weighted to proportionately contribute to both the phrases, corresponding to the \textit{shoes} and the \textit{pants}, separately. A phrase-conditioned region representation is then obtained as the assignment-weighted average of patch embeddings, where $\ell_2(\cdot)$ denotes $\ell_2$-normalisation:
\begin{equation}
\hat{\bm{z}}_{i,p} =
\ell_2 \left( \sum_{k=1}^{K} a_{i,k,p} \bm{z}_{i,k} \right)
\label{eq:agg_patch}
\end{equation}

\noindent \textbf{Coverage Regularisation.} Since the softmax in Eq.~\ref{eq:patch_assign} is applied over phrases for each patch, the model may converge to a degenerate solution in which most patches assign their probability mass to a single dominant phrase. In this case, that phrase receives the majority of the gradient signal, while other valid phrases obtain negligible assignment weights and thus weak supervision. This collapse prevents meaningful phrase-conditioned region learning. To mitigate this, we define the expected coverage of phrase $p$ as $c_{i,p} = \sum_k a_{i,k,p}$ and maximize its average over valid phrases, ensuring that each phrase attracts non-trivial patch mass and receives sufficient training signal. We define coverage loss, $\mathcal{L}_{cov}$, which ensures a stable collapse-free training. Here $B$ is the batch size, $m_{i,p} \in \{0,1\}$ indicates if phrase $p$ is valid for sample $i$, and $n_i = \sum_p m_{i,p}$.
\begin{equation}
\mathcal{L}_{\text{cov}} =
-\frac{1}{B} \sum_{i=1}^{B}
\frac{1}{n_i}
\sum_{p=1}^{P}
m_{i,p}\, c_{i,p}
\label{eq:cov_loss}
\end{equation}

\noindent \textbf{Per-Phrase Cross-Image Contrastive Loss.}
The main supervision arises from a cross-image contrastive loss applied independently for each phrase. 
This objective encourages phrase-conditioned region representations to be discriminative across identities. 
For a given phrase $p$, the aggregated region vector $\hat{\bm{z}}_{i,p}$ from image $i$ is contrasted against all phrase embeddings $\{\bm{h}_{j,p}\}_{j=1}^{B}$ in the batch. 
Regions originating from images that share the same identity are treated as positives, while those from different identities are treated as negatives. 
This contrastive supervision encourages the aggregated region representation to capture identity-consistent visual evidence associated with the phrase, rather than diffuse or background features.

To consistently satisfy this objective across many images with different people, poses, and backgrounds, the model must aggregate patches that correspond to the visual attribute described by the phrase (e.g., the torso region for a \textit{red hoodie}). 
A diffuse average over all patches would include background, floor, and unrelated clothing regions, making the aggregated representation less discriminative across identities and thus increasing the loss. 
Consequently, the cross-image objective implicitly encourages spatially focused phrase-region alignment.

For phrase position $p$, we stack the aggregated region vectors $\hat{\bm{z}}_{i,p}$ and phrase embeddings $\bm{h}_{i,p}$ across the batch to form matrices $\hat{\bm{Z}}_p$ and $\bm{H}_p \in \mathbb{R}^{B \times D}$, respectively, and compute the similarity matrix: 
$\bm{Q}_p = \hat{\bm{Z}}_p \bm{H}_p^\top$,
where $\bm{Q}_p[i,j] = \langle \hat{\bm{z}}_{i,p}, \bm{h}_{j,p} \rangle$ measures the similarity between the region representation extracted from image $i$ using its $p$-th phrase and the $p$-th phrase embedding of image $j$. 
Note that phrases at position $p$ are not required to correspond to the same semantic attribute across images; identity labels determine positive and negative pairs in the contrastive objective.

Images sharing the same identity ($y_i = y_j$) are treated as positives.
TAL is applied bidirectionally:
\begin{equation}
\mathcal{L}_p =
\frac{1}{2B}
\sum_{i=1}^{B}
m_{i,p}\,\hat{y}_i
\Big[
\ell_{\text{TAL}}(\bm{Q}_p[i,:]) +
\ell_{\text{TAL}}(\bm{Q}_p[:,i])
\Big]
\label{eq:per_phrase_loss}
\end{equation}
where $\hat{y}_i \in [0,1]$ is the per-sample confidence weight inherited from the RDE base framework~\cite{qin2024noisy}, which down-weights image-text pairs flagged as noisy correspondences so that unreliable samples contribute less to the part objective.
Phrase losses are aggregated using frequency-based weights
$w_p = f_p / \sum_{p'} f_{p'}$,
where $f_p = \sum_i m_{i,p}$.
The total part contrastive loss is: 
\begin{equation}
\mathcal{L}_{\text{part}} =
\sum_{p:\,f_p>0} w_p\, \mathcal{L}_p
\label{eq:part_contrastive}
\end{equation}

\noindent \textbf{Locality Induction.}
The cross-image objective encourages phrase-conditioned region representations to be discriminative across identities.
Diffuse aggregation over unrelated patches reduces cross-image separability and increases contrastive loss. As a result, the model is incentivized to concentrate phrase evidence into spatially consistent subsets of patches. 

% ----------------------------
\noindent \textbf{Combined Training Objective.}
\label{sec:combined_loss}
% ----------------------------
%
The final objective is the combination of global base retrieval loss along with the new part loss, which has two components: total part contrastive loss and the coverage loss, as described above.
\begin{equation}
\mathcal{L} =
\mathcal{L}_{\text{base}}
+
\lambda_{\text{part}}\, r(e)
\left(
\mathcal{L}_{\text{part}}
+
\lambda_{\text{cov}} \mathcal{L}_{\text{cov}}
\right),
\label{eq:full_loss}
\end{equation}
where $r(e) = \min(1, (e+1)/E_{\text{warm}})$ linearly warms up the part loss during early $E_{\text{warm}}$  training epochs.
This prevents unstable gradients before global representations are sufficiently formed.

\section{Quantitative Interpretability Framework}
Previous methods on interpretability of person TI-ReID models \cite{park2024plot, kim2026dico} introduce a slot-attention module between the text and vision encoders to learn shared ``concepts'' within a set of slots. In our view, this approach leverages the inherent object-centric disentanglement property of slot-attention \cite{locatello2020object}, where the emergence of disentangled representations is part of the core architectural design. As such, these works rely primarily on the inductive bias of the slot-attention mechanism and visualize the learned slot weights to infer qualitative interpretability. However, no explicit quantitative metric is introduced to measure interpretability in a distinct and controlled manner. Additionally, incorporating slot-attention substantially increases model complexity. For example, the ICL \cite{qin2025human} method with approximately 153 million parameters achieves overall better performance than PLOT \cite{park2024plot}, which has approximately 292 million parameters. This suggests that interpretability should not necessarily be tied to parameter scaling.

In our work, we propose an interpretability framework that produces qualitative explanations comparable to prior methods without introducing an increase in parameters. We operate under the same parameter budget as ICL~\cite{qin2025human}. More importantly, we introduce a quantitative framework for measuring interpretability performance. We define the fundamental units of explanation as \emph{interpretable units}, and evaluate a method's ability to extract these units and quantify their importance in explaining retrieval outcomes. We rank the interpretable units (parts) according to their relative contribution to the retrieval task. We then mask the highest-ranked parts and measure the resulting drop in retrieval performance. This masking-and-evaluation strategy is conceptually similar to the stratified inpainting approach proposed in \cite{cohen2025meaningful}. We define interpretability strength such that larger performance drops correspond to more meaningful interpretable units. The goal of our evaluation is to determine whether these interpretable units, are causally responsible for the retrieval performance.

% ----------------------------
\noindent \textbf{Counterfactual Region-Removal Evaluation.}
\label{sec:eval_protocol}
% ----------------------------
%
The central idea is to test \emph{causal faithfulness}: if a highlighted region is genuinely responsible for a high query--image similarity, removing that region should reduce the similarity and degrade retrieval performance. Similar approaches have been introduced in the context of interpretability and explainability of transformers~\cite{chefer2021generic} and in this study~\cite{cohen2025meaningful} where the paper uses perturbation-based methods for interpreting and explaining decisions made by a deep neural network.

%%%%%%%
\noindent \textbf{Retrieval formulation.}
Let $\mathcal{Q}=\{t_i\}_{i=1}^{N_q}$ denote the set of query texts and $\mathcal{G}=\{x_j\}_{j=1}^{N_g}$ the gallery images. 
A retrieval model produces embedding representations for both modalities, yielding a similarity matrix \(S_{ij} = \langle \mathbf{q}_i , \mathbf{g}_j \rangle\).
Where $\mathbf{q}_i$ and $\mathbf{g}_j$ are normalized query and gallery embeddings. For each query $t_i$, the baseline top-ranked gallery image is: 
\begin{equation}
j_i = \arg\max_{j} S_{ij}.
\label{eq:top_rank_selection}
\end{equation}

%%%%%%%%%
\noindent \textbf{Phrase-conditioned relevance and a two-stage part-local mask.}
Given a query $t_i$, we perform the retrieval and then use the query and retrieved image to obtain a set of phrase-level parts $\mathcal{L}_i=\{\ell_{i,m}\}_{m=1}^{P_i}$. These phrases can be obtained in several ways, ranging from simple syntactic decompositions of the text query to user-provided phrase annotations. In our implementation, a multimodal LLM (MLLM) is employed to extract semantically meaningful phrase-level descriptions from the query and the retrieved image.
For each phrase $\ell_{i,m}$ and retrieved image $x_{j_i}$, we compute the relevance map $R_{i,m}\in\mathbb{R}^{H\times W}$, as described by Eq.~\ref{eq:phrase_to_patch_sim} (upsampled to image resolution), where larger values indicate higher phrase-conditioned relevance.

% --- Thresholded region + top-p removal inside region ---
We convert $R_{i,m}$ into a binary perturbation mask using a two-stage selection that separates
(i) \emph{where} the part is localized and (ii) \emph{how much} of that localized support is removed.
Let $\Omega=\{1,\dots,H\}\times\{1,\dots,W\}$ denote pixel coordinates and let
$R_{i,m}(\mathbf{u})$ be the relevance score at pixel $\mathbf{u}\in\Omega$.

\noindent \textbf{Stage 1: thresholded part region.}
We first define the \emph{part region} by thresholding:
\begin{equation}
\mathcal{I}^{\mathrm{reg}}_{i,m}(\alpha)
\;=\;
\{\mathbf{u}\in\Omega \;:\; R_{i,m}(\mathbf{u}) \ge \alpha\},
\label{eq:region_def_thr}
\end{equation}
where $\alpha\in[0,1]$ controls what is the strength of the region we are selecting (larger $\alpha$ yields a smaller, higher-confidence region). $\alpha = 1$ means that no pixels are selected to be masked, which makes sense because in a normalized part-mask, none of the pixels exceed 1. And conversely, for $\alpha = 0$, all the pixels exceed this threshold and thus are subject to being selected as the part-region. $p$ fraction of those selected pixels are then used to mask out the relevant part (region) in the image. 

\noindent \textbf{Stage 2: top-$p$ removal within the region.}
Let $\mathrm{Top}(\mathcal{S},k)$ return the subset of $\mathcal{S}$ corresponding to the $k$ largest values of $R_{i,m}(\mathbf{u})$ over $\mathbf{u}\in\mathcal{S}$.
We then select the top-$p$ fraction \emph{within} the region:
\begin{equation}
\mathcal{I}^{\mathrm{mask}}_{i,m}(p,\alpha)
\;=\;
\mathrm{Top}\!\Big(\mathcal{I}^{\mathrm{reg}}_{i,m}(\alpha),\;
\big\lceil p\,|\mathcal{I}^{\mathrm{reg}}_{i,m}(\alpha)|\big\rceil\Big),
\label{eq:mask_def_thr}
\end{equation}
with $p\in[0,1]$ controlling the perturbation strength. If we set $p=0$ that would mean that none of the pixels are going to be perturbed, thus the gallery image remains unchanged, and hence, our baseline is maintained. And conversely, if $p=1$, all the pixels within the selected region are perturbed, and we see the maximum drop in the retrieval performance.
The binary mask $M_{i,m}(p,\alpha)\in\{0,1\}^{H\times W}$ is defined as: 
$M_{i,m}(p,\alpha)(\mathbf{u})=1$ if $\mathbf{u}\in\mathcal{I}^{\mathrm{mask}}_{i,m}(p,\alpha)$ and $0$ otherwise.
This construction makes $p$ interpretable as ``\% removed \emph{in the thresholded part region}'', while $\alpha$ determines which pixels are deemed activated enough to belong to the part region.

\noindent \textbf{Counterfactual gallery perturbation.}
A perturbed gallery image is formed by removing the selected pixels (we set them to $0$ in the normalized image space):
\begin{equation}
\tilde{x}_{i,m}(p,\alpha)
\;=\;
\mathrm{Perturb}\!\big(x_{j_i},\, M_{i,m}(p,\alpha)\big).
\label{eq:perturb_img}
\end{equation}
We re-encode this perturbed image to obtain $\tilde{\mathbf{g}}_{i,m}$ and compute the counterfactual similarity, and then use that to compute the similarity drop \(\Delta s_{i,m}(p,\alpha)\): 
\begin{equation}
\tilde{S}^{(m)}_{i}(p,\alpha)
=
\langle \mathbf{q}_i,\tilde{\mathbf{g}}_{i,m}\rangle.
\label{eq:sim_after_single}
\end{equation}
\begin{equation}
\Delta s_{i,m}(p,\alpha)
=
S_{i j_i} - \tilde{S}^{(m)}_{i}(p,\alpha).
\label{eq:delta_sim}
\end{equation}

\begin{figure}[!h]
\centering
\includegraphics[width=0.8\linewidth,trim={8pt 10pt 6pt 8pt},clip]{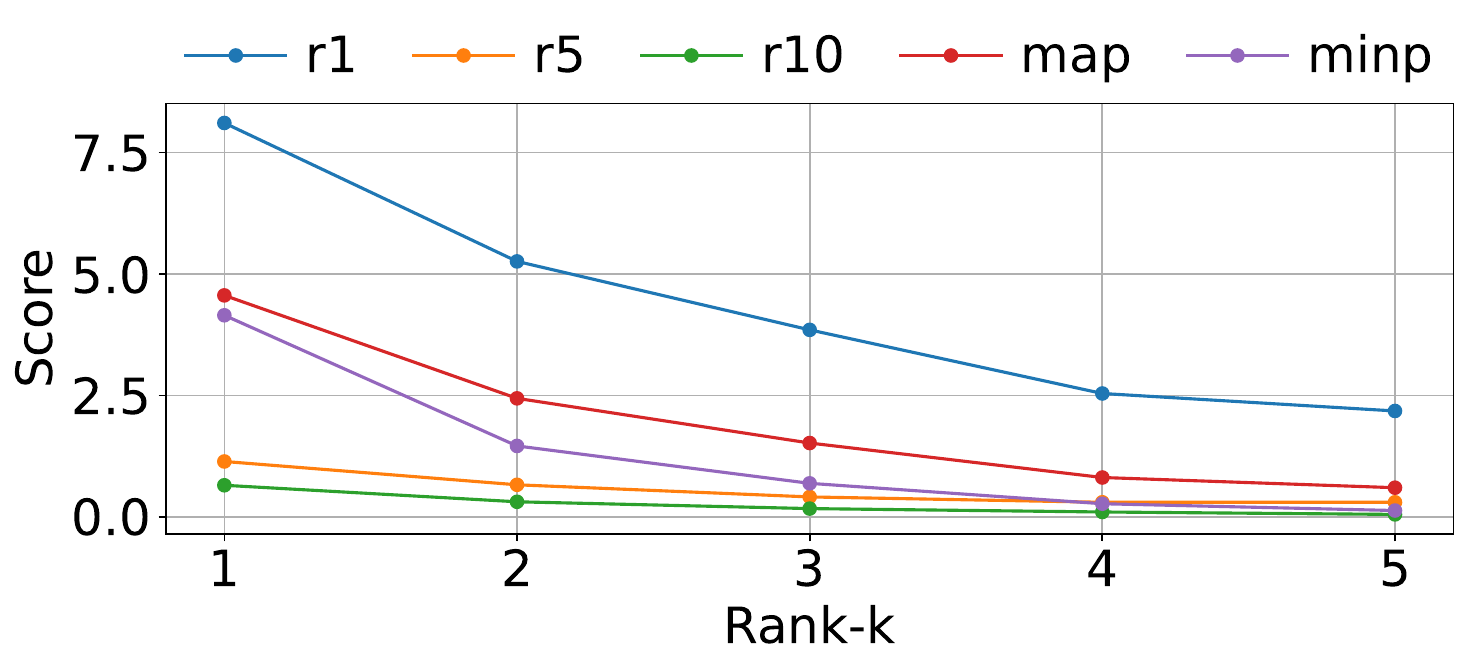}
\caption{\textbf{Per-rank counterfactual drop (CUHK-PEDES).}
Marginal relative drop ($\Delta\%$) in each retrieval metric when \emph{only} the $k$-th most discriminative part is removed. All metrics decrease monotonically with $k$: the rank-1 part carries the largest causal weight, and later, more discriminative parts contribute progressively less, so rank-1 removal faithfully summarises ranking.
\vspace{-0.1cm}}
\label{fig:rankk}
\end{figure}

\paragraph{Rank-1 influential part removal.}
The most influential phrase is defined as \(m_i^\star\), which is the most important interpretable unit among all the interpretable units identified by the model: 
\begin{equation}
m_i^\star
=
\arg\max_{m} \Delta s_{i,m}(p,\alpha).
\label{eq:rank1_part}
\end{equation}
The single most discriminative part is removed because this isolates the strongest causal signal. A rank-$k$ ablation on CUHK-PEDES (Fig.~\ref{fig:rankk}) shows every metric decreasing monotonically as $k$ grows: the top part carries the largest causal weight and subsequent parts contribute progressively less. Cumulative top-$k$ removal therefore yields a trivially increasing curve, so rank-1 is a faithful summary of the full ranking rather than a cherry-picked operating point.

\noindent \textbf{Similarity-matrix update.}
Recall from Eq.\ref{eq:top_rank_selection} that $j_i = \arg\max_j S_{ij}$ denotes the index of the baseline top-1 retrieved gallery image for query $t_i$.
After identifying the most influential phrase $m_i^\star$, we compute the perturbed similarity
$\tilde{S}^{(m_i^\star)}_{i}(p,\alpha)$ for the pair $(t_i, x_{j_i})$. We then construct a counterfactual similarity matrix by updating only the attacked similarity entry corresponding to the baseline top-1 pair:
\begin{equation}
S^{\mathrm{cf}}_{ij}
=
\begin{cases}
\tilde{S}^{(m_i^\star)}_{i}(p,\alpha), & (i,j)=(i,j_i), \\
S_{ij}, & \text{otherwise}.
\end{cases}
\label{eq:sim_matrix_update}
\end{equation}
In other words, only the similarity between query $t_i$ and its originally retrieved image $x_{j_i}$ is replaced with the counterfactual similarity, while all other query-gallery similarities remain unchanged. This single-cell update isolates the causal effect of removing the explanation region while leaving all other similarities unchanged.

\noindent \textbf{Metric drops.}
Let $\mathcal{M}$ denote any retrieval metric (R@K mAP, mINP).  
We compute \(\Delta \mathcal{M}(p,\alpha) = \mathcal{M}(S) - \mathcal{M}(S^{\mathrm{cf}}(p,\alpha))\) and the relative drop \(\Delta \mathcal{M}\%\): 

\begin{equation}
\Delta\mathcal{M}\% =
100\frac{\Delta\mathcal{M}}{\mathcal{M}(S)}.
\label{eq:metric_drop}
\end{equation}

%%%%%%%%%%%%%%%%%%%%%%%%%%%%%%%%%%
\section{Results and Discussion}

\subsection{Experimental Methodology}
\noindent \textbf{Datasets.} The datasets used here are CUHK-PEDES dataset \cite{li2017person} which contains 40,206 samples with 13,003 unique individuals, and each image sample has two textual descriptions. Then we have the ICFG-PEDES dataset \cite{ding2021semantically} which contains a total of 54,522 images and 54,522 text descriptions across 4,102 unique person identities with one caption per image sample. RSTPReid dataset \cite{rstpreid}, which contains 20,505 images of 4,101 persons, with each image annotated with two captions. They contain image-text pairs, the images are of pedestrians, hence making it appropriate for the person ReID task.

\noindent \textbf{Baseline methods.} We compare against recent TI-ReID methods spanning both retrieval and interpretability paradigms. IRRA~\cite{jiang2023cross}, APTM~\cite{yang2023towards}, NAM~\cite{tan2024harnessing}, and ICL~\cite{qin2025human} are retrieval-oriented: they optimize global image-text similarity and produce no part-level explanations whatsoever. Moreover, PLOT~\cite{park2024plot} and DiCo~\cite{kim2026dico} are the only prior interpretable TI-ReID methods. The code for DiCo was not publicly available, preventing direct benchmarking. However, based on our assessment, both methods rely on a similar slot-based formulation. 

% ----------------------------
% Table: sensitivity
% ----------------------------
\subsection{Comparison with State-of-the-Art Methods}
Tab.\ref{tab:ppim_sensitivity} reports the relative degradation in retrieval metrics when the most influential part (interpretable unit) from the top gallery image for each query is removed using the proposed counterfactual masking protocol.
Across all datasets, proposed patch-phrase interaction module (PPIM) consistently increases the sensitivity of our model to localized perturbations.
In other words, when PPIM is used, removing the most explanatory region produces larger drops in retrieval performance.
For instance, on CUHK-PEDES, the R@1 degradation increases from $6.13\%$ to $8.72\%$, while on RSTPReID, the drop more than doubles from $7.75\%$ to $11.57\%$.
Similar trends are observed across all ranking metrics and datasets. We also establish a new interpretability performance state-of-the-art for all datasets.

\begin{table}[t]
\vspace{-0.3cm}
\centering
\setlength{\tabcolsep}{3pt}
\renewcommand{\arraystretch}{0.9}
\resizebox{\linewidth}{!}{
\begin{tabular}{l|c|ccccc|cc}
\hline
\multirow{2}{*}{\textbf{Dataset}} & \multirow{2}{*}{\textbf{Method}}
& \multicolumn{5}{c|}{\textbf{$\Delta\mathcal{M}\%$}\,$\uparrow$}
& \multirow{2}{*}{\textbf{Params}} \\% (M)} $\downarrow$} \\
\cline{3-7}
& & \textbf{R@1} & \textbf{R@5} & \textbf{R@10} & \textbf{mAP} & \textbf{mINP} &%& Trainable (M) 
\\
\hline
& Baseline & 6.13 & 1.11 & 0.45 & 3.33 & 2.93 & 153 \\
CUHK-PEDES & PLOT & 2.81 & 0.14 & 0.05 & 1.01 & 0.45 & 292 \\
& $+$PPIM & \textbf{8.72} & \textbf{1.71} & \textbf{0.74} & \textbf{4.76} & \textbf{4.46} & 153 \\
\hline
& Baseline & 3.97 & 0.92 & 0.35 & 0.78 & 0.03 & 153 \\
ICFG-PEDES & PLOT & 2.15 & 0.43 & 0.13 & 0.33 & 0.00 & 292 \\
& $+$PPIM & \textbf{6.27} & \textbf{1.85} & \textbf{0.92} & \textbf{2.11} & \textbf{0.31} & 153 \\
\hline
& Baseline & 7.75 & 0.75 & 0.61 & 2.24 & 0.56 & 153 \\
RSTPReID & PLOT & 4.21 & 0.43 & 0.06 & 1.32 & 0.40 & 292 \\
& $+$PPIM & \textbf{11.57} & \textbf{2.46} & \textbf{0.72} & \textbf{4.32} & \textbf{0.80} & 153 \\
\hline
\end{tabular}
}
\caption{\textbf{Perturbation sensitivity under PPIM-based counterfactual masking (rank-1 part).}
We report relative metric drops ($\Delta\mathcal{M}\%$; Eq.\ref{eq:metric_drop}, $\alpha=0.3$, $p=0.1$). Larger drops indicate stronger causal reliance on localized phrase-aligned regions. Across all three benchmarks, $+$PPIM yields the largest degradation on every metric, substantially exceeding both the $-$PPIM baseline and PLOT while using $1.9\times$ fewer parameters (153M vs.\ 292M). On CUHK-PEDES, $+$PPIM raises the R@1 drop to $8.72\%$ ($3.1\times$ PLOT), and on RSTPReID it reaches $11.57\%$ ($2.7\times$ PLOT), confirming that PPIM grounds retrieval decisions in spatially concentrated, semantically meaningful evidence.
\vspace{-0.2cm}}
\label{tab:ppim_sensitivity}
\end{table}

This behavior indicates that the representations learned with PPIM rely more strongly on spatially localized visual evidence aligned with the textual phrases.
Consequently, masking the region identified as explanatory disrupts the cross-modal similarity more significantly.
In contrast, the baseline model without PPIM exhibits smaller metric drops, suggesting that its similarity scores depend less on well-localized semantic regions and more on diffuse global features.
Overall, these results provide quantitative evidence that PPIM strengthens the causal alignment between phrase-level visual evidence and retrieval decisions, leading to explanations that better reflect the model's decision-making process.

\begin{figure}[!b]
    \centering
    \begin{subfigure}{\linewidth}
        \centering
        \begin{overpic}[width=\linewidth,trim={0 7pt 0 10pt},clip]{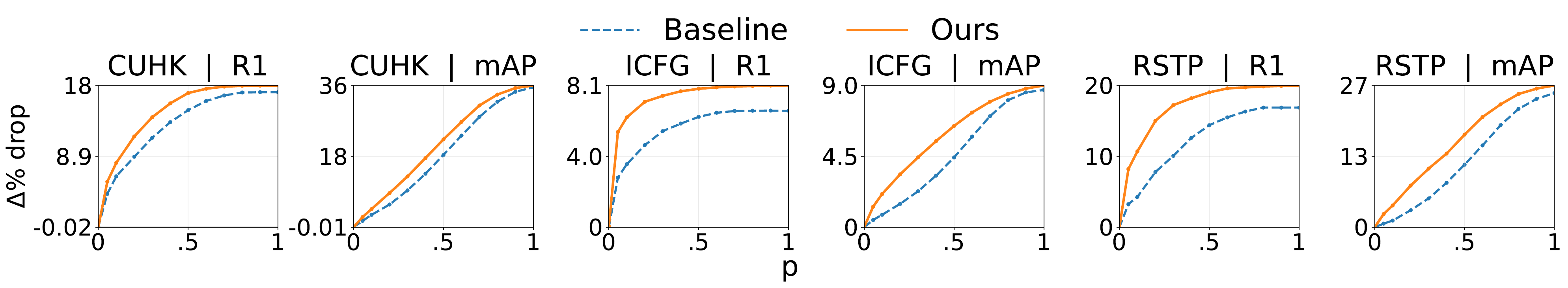}
            \put(50,90){\makebox(0,0)
            }
        \end{overpic}
        \caption{Sensitivity to perturbation strength.}
        \label{fig:p_ablation}
    \end{subfigure}
    \vspace{0.2em}
    \begin{subfigure}{\linewidth}
        \centering
        \begin{overpic}[width=\linewidth,trim={0 6pt 0 10pt},clip]{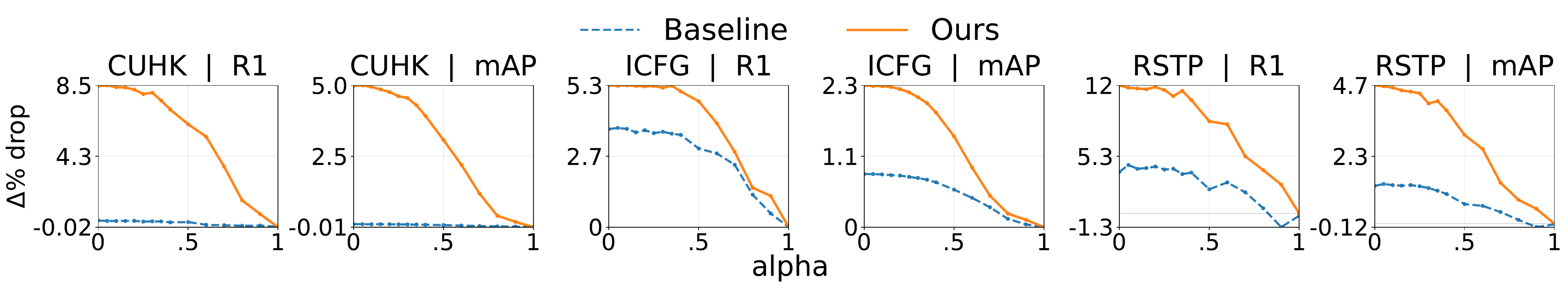}
            \put(50,90){\makebox(0,0)
            }
        \end{overpic}
        \caption{Sensitivity to thresholded relevance masking.}
        \label{fig:alpha_ablation}
    \end{subfigure}
    \caption{\textbf{Sensitivity analysis of relevance-based masking.}
    \textbf{(a)} Sensitivity to perturbation strength $p$ (fixed $\alpha=0.3$) and \textbf{(b)} to the relevance threshold $\alpha$ (fixed $p=0.1$), relative drops ($\Delta\%$) in R@1 and mAP for the most influential phrase region (Eq.~\ref{eq:rank1_part}). 
    \vspace{-0.3cm}
    }
\end{figure}

Fig.\ref{fig:p_ablation} shows the effect of increasing the perturbation strength $p$, where the y-axis reports the relative metric drop ($\Delta\%$) after masking the selected interpretable regions.
Across all datasets and metrics, the performance degradation increases smoothly as larger fractions of the highlighted regions are removed.
Notably, our method consistently exhibits larger metric drops than the baseline.
This indicates that the regions identified by our part-based explanations correspond to visual evidence that the model genuinely relies on for cross-modal matching.
As progressively larger portions of these regions are removed, the similarity between the query and the retrieved image decreases, leading to a monotonic increase in retrieval degradation.

Fig.\ref{fig:alpha_ablation} analyzes the effect of the relevance threshold $\alpha$, which controls the spatial extent of the selected regions.
Smaller values of $\alpha$ include larger portions of the activation map, resulting in stronger perturbations and larger metric drops.
As $\alpha$ increases, the selected regions become progressively more localized around the most confident activations, reducing the amount of removed evidence and therefore decreasing the observed degradation.
Importantly, our method consistently produces larger drops than the baseline across datasets, suggesting that the identified regions correspond more closely to the visual evidence driving retrieval decisions.
The smooth and stable trends observed across both parameters indicate that retrieval relies on spatially distributed semantic regions rather than isolated pixels.
If the highlighted regions were not aligned with the model's decision process, masking them would produce negligible or unstable changes in retrieval metrics.
Instead, the consistent degradation across datasets supports that the proposed part-based explanations capture causally relevant regions influencing cross-modal similarity.

\begin{figure}[b]
    \centering
    \includegraphics[width=0.92\linewidth,trim={24pt 13pt 20pt 11pt},clip]{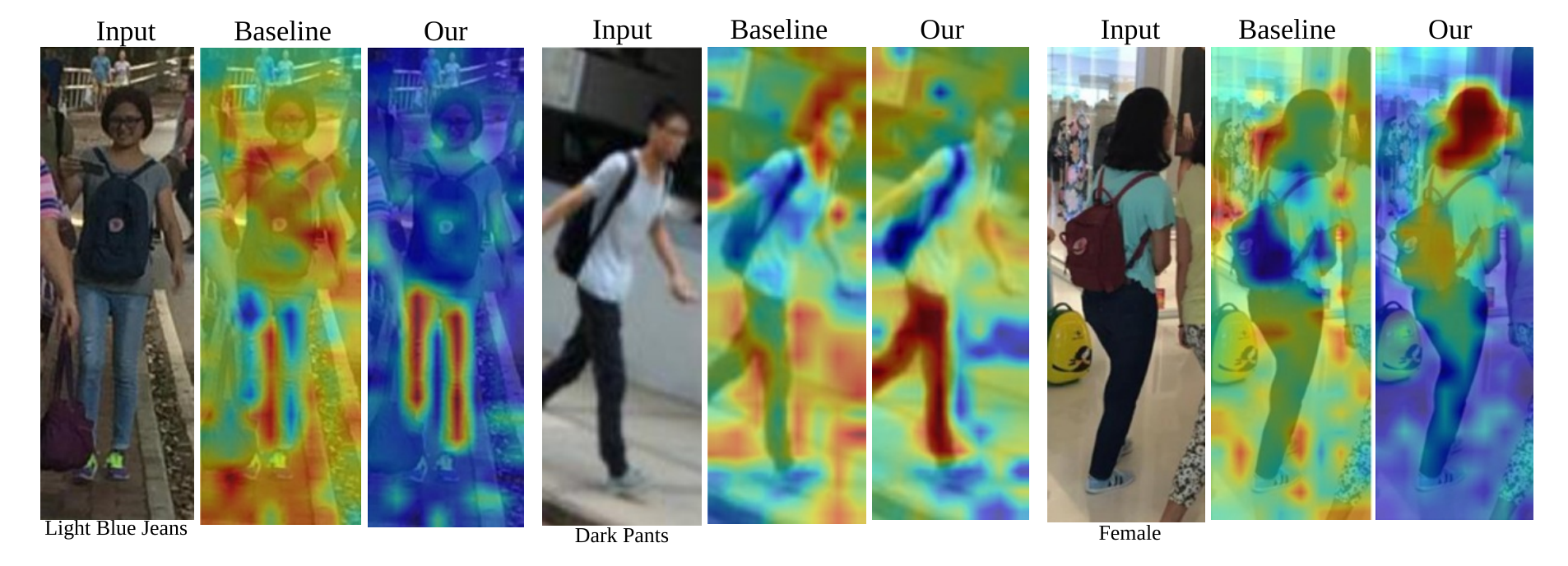}
    \caption{%
    \textbf{Qualitative phrase-conditioned heatmaps.} Our method gives better part-localization as compared to the baseline.
    \vspace{-0.9cm}
    }
    \label{fig:qual_heatmaps}
\end{figure}
Fig.\ref{fig:qual_heatmaps} presents qualitative comparisons of phrase-conditioned relevance maps produced by the baseline model and by our method trained with the proposed patch-phrase interaction module (PPIM). 
Across different query phrases (e.g., \emph{light blue shirt}, \emph{dark pants}, and \emph{female}), the baseline model produces relatively diffuse activations that spread across multiple body regions and background areas. 
In contrast, our method yields significantly sharper and more spatially localized responses that align closely with the semantic regions referenced by the phrase. This behavior indicates that the proposed part loss encourages the model to associate textual phrases with semantically meaningful visual regions rather than relying on coarse global cues. 
Consequently, the learned representations exhibit stronger phrase-to-region grounding, where distinct attributes activate the corresponding body parts more precisely. 
%
%%%%%%%%%%%%%%%%%%%%%%%%%%%%%%%%%%%
Tab.\ref{tab_main} shows that we achieve competitive retrieval performance across all the benchmarks, in some cases even higher than the current SOTA. Further proving that our PPIM module not only helped achieve SOTA interpretability performance but also maintained or exceeded the retrieval performance. 

\begin{table}[h]
\vspace{-0.16cm}
\centering 
\setlength{\abovecaptionskip}{0.1cm}
\arrayrulecolor{black} 
\setlength\tabcolsep{1.5pt}
\resizebox{1\linewidth}{!}{
\begin{tabular}{l|cc|c|cc|cc|cc}\hline\hline
\rowcolor{gray!25}   &    &    &   & \multicolumn{2}{c|}{{\textbf{CUHK-PEDES}}} & \multicolumn{2}{c|}{{\textbf{ICFG-PEDES}}} & \multicolumn{2}{c}{{\textbf{RSTPReid}}}                          \\
\rowcolor{gray!25} \textbf{Methods} & \textbf{Image Enc.}&\textbf{Text Enc.} & \textbf{Interp.} & \textbf{Rank-1} & \textbf{mAP}
& \textbf{Rank-1} & \textbf{mAP}
& \textbf{Rank-1} & \textbf{mAP} \\\hline
IRRA~\cite{jiang2023cross} \textit{(cvpr'23)} & CLIP-ViT & CLIP-X. & $\times$ & 74.05 & 66.57 & 64.37 & 38.85 & 61.90 & 48.08 \\
APTM~\cite{yang2023towards} \textit{(mm'24)} & Swin-B & BERT & $\times$ & 76.53 & 66.91 & 68.51 & 41.22 & 67.50 & 52.56 \\
NAM$^\natural$~\cite{tan2024harnessing} \textit{(cvpr'24)} & CLIP-ViT & CLIP-X. & $\times$ & 77.47 & 69.43 & 66.76 & 41.45 & 67.15 & 52.00 \\
PLOT~\cite{park2024plot} \textit{(eccv'24)} & CLIP-ViT & CLIP-X. & $\checkmark$ & 75.28 & - & 65.76 & - & 61.80 & - \\
ICL~\cite{qin2025human} \textit{(cvpr'25)} & CLIP-ViT & CLIP-X. & $\times$ & 78.18 & 69.58 & 69.22 & 42.34 & 70.00 & \textbf{54.16} \\
DiCO~\cite{kim2026dico} \textit{(n-computing'26)} & CLIP-ViT & CLIP-X. & $\checkmark$ & 77.21 & - & 67.81 & - & 67.84 & - \\
\hline
\textbf{\method} & CLIP-ViT & CLIP-X. & $\checkmark$ & \textbf{78.28} & 69.26 & \textbf{69.92} & 42.03 & \textbf{71.05} & 52.76 \\
\hline\hline
\end{tabular} 
} 
\caption{Performance on the three coarse-grained benchmarks. Detailed results are in the supplementary materials.
\vspace{-0.33cm}
} 
\label{tab_main}
\end{table}

%%%%%%%%
\subsection{Robustness to Phrase Annotation Noise}
Because phrases are generated automatically rather than annotated by humans, a concern is whether annotation errors degrade supervision. We analyze annotation quality, sensitivity to the generation source, and
whether the measured drops are causal.

\noindent\textbf{Annotation quality.}
Our generated annotations have a fault rate of $16.5$--$17.5\%$ over sample$\times$category pairs, leaving ${\sim}10$ valid phrases per sample as supervision, across the three datasets. \method{} is robust to these faults. \textbf{(i)} Faulty fields never impact the loss: the dataloader sets $m_{i,p}{=}0$ for invalid fields, so they contribute zero to the contrastive (Eq.\ref{eq:per_phrase_loss}) and coverage (Eq.\ref{eq:cov_loss}) terms. \textbf{(ii)} Errors are scattered across samples, whereas correct concepts recur across images of the same identity. For an identity with $k$ correctly-annotated images and a noisy annotation, the $k$ gradients pulling the aggregated region toward a consistent phrase embedding dominate the lone outlier. Repeated concepts are reinforced while isolated errors wash out.

\noindent\textbf{Effect of diverse phrase-annotations.}
Tab.\ref{tab:mllm} shows that across phrase-generation
configurations: two prompts conditioned on caption and image (P1, P2) and one using the caption only (P3). Metric stays stable across a diverse set of phrase-annotations. Showing our method works on varied annotations.

\begin{table}[!ht]
\centering
    \centering
    \setlength{\tabcolsep}{4pt}%
    \resizebox{0.59\linewidth}{!}{%
    \begin{tabular}{lccc}
    \toprule
    \textbf{Prompt} & \textbf{Caption} & \textbf{Image} & \textbf{$\Delta$R@1 (\%)} \\
    \midrule
    P1 & \checkmark & \checkmark & 8.11 \\
    P2 & \checkmark & \checkmark & 8.26 \\
    P3 & \checkmark & --         & 8.95 \\
    \bottomrule
    \end{tabular}}
    \caption{\textbf{Effects of varied phrase-annotations} (CUHK-PEDES). 
    }
    \label{tab:mllm}
\end{table}

\noindent\textbf{Removal is causal, not an artifact of masking (randomized control experiment).}
As a control, we remove the same number of pixels at uniformly random
locations on CUHK-PEDES: $\Delta$R@1$=5.90\%$, $\Delta$mAP$=3.02\%$,
versus $8.72\%$ and $4.76\%$ for \method{}-guided removal. The degradation
tracks the regions \method{} identifies rather than the act of masking.

\section{Conclusion}

This paper addressed a key limitation of TI-ReID systems: although recent VLMs achieve strong retrieval performance, their predictions remain largely opaque and unexplained. The model cannot know the part of its representation that is responsible for a given match. We introduced \method{}, a CLIP-based method that promotes explicit phrase–region grounding by encouraging localized patch–phrase alignment during training.  Our PPIM provides lightweight concept-level supervision that associates textual phrases with semantically meaningful visual regions without relying on slot-based decomposition. Beyond architectural design, we proposed a quantitative interpretability protocol for TI-ReID based on counterfactual region masking. It enables the systematic evaluation of whether highlighted regions are causally responsible for predicted retrievals.  Experiments across standard benchmarks indicate that \method{} produces spatially grounded explanations while maintaining competitive retrieval accuracy, showing that interpretability and retrieval accuracy are not conflicting objectives.

\noindent\textbf{Supplementary material.} Code, sample phrase annotations, including input prompts, extended related work, and extended results are in the supplementary material.

{
    \small
    \bibliographystyle{ieeenat_fullname}
    \bibliography{main}

@inproceedings{qin2025human,
  title        = {Human-centered Interactive Learning via MLLMs for Text-to-Image Person Re-identification},
  author       = {Y. Qin and
                  C. Chen and
                  Z. Fu and
                  et al.},
  booktitle    = {CVPR},
  year         = {2025}
}

@inproceedings{bai2025chat,
  title        = {Chat-based Person Retrieval via Dialogue-Refined Cross-Modal Alignment},
  author       = {Y. Bai and
                  Y. Ji and
                  M. Cao and
                  et al.},
  booktitle    = {CVPR},
  year         = {2025}
}

@inproceedings{li2017person,
  title        = {Person Search with Natural Language Description},
  author       = {S. Li and
                  T. Xiao and
                  H. Li and
                  et al.},
  booktitle    = {CVPR},
  year         = {2017}
}

@inproceedings{zhang2018deep,
  title        = {Deep Cross-Modal Projection Learning for Image-Text Matching},
  author       = {Y. Zhang and
                  H. Lu},
  booktitle    = {ECCV},
  year         = {2018}
}

@inproceedings{jing2020pose,
  title        = {Pose-guided Multi-granularity Attention Network for Text-based Person Search},
  author       = {Y. Jing and
                  C. Si and
                  J. Wang and
                  et al.},
  booktitle    = {AAAI},
  year         = {2020}
}

@inproceedings{wang2020vitaa,
  title        = {Vitaa: Visual-Textual Attributes Alignment in Person Search by Natural Language},
  author       = {Z. Wang and
                  Z. Fang and
                  J. Wang and
                  et al.},
  booktitle    = {ECCV},
  year         = {2020},
}

@article{ding2021semantically,
  title        = {Semantically Self-Aligned Network for Text-to-Image Part-Aware Person Re-identification},
  author       = {Z. Ding and
                  C. Ding and
                  Z. Shao and
                  et al.},
  journal      = {CoRR},
  volume       = {arXiv:2107.12666},
  year         = {2021}
}

@inproceedings{park2024plot,
  title        = {Plot: Text-based Person Search with Part Slot Attention for Corresponding Part Discovery},
  author       = {J. Park and
                  D. Kim and
                  B. Jeong and
                  et al.},
  booktitle    = {ECCV},
  year         = {2024},
  organization = {Springer}
}

@article{locatello2020object,
  title        = {Object-Centric Learning with Slot Attention},
  author       = {F. Locatello and
                  D. Weissenborn and
                  T. Unterthiner and
                  et al.},
  journal      = {Adv. Neural Inf. Process. Syst.},
  year         = {2020}
}

@inproceedings{cohen2025meaningful,
  title        = {A Meaningful Perturbation Metric for Evaluating Explainability Methods},
  author       = {D. Cohen and
                  H. Chefer and
                  L. Wolf},
  booktitle    = {SCIA},
  year         = {2025},
}

@article{gao2021contextual,
  title        = {Contextual Non-Local Alignment over Full-Scale Representation for Text-based Person Search},
  author       = {C. Gao and
                  G. Cai and
                  X. Jiang and
                  et al.},
  journal      = {CoRR},
  volume       = {arXiv:2101.03036},
  year         = {2021}
}

@inproceedings{radford2021learning,
  title        = {Learning Transferable Visual Models from Natural Language Supervision},
  author       = {A. Radford and
                  J. Kim and
                  C. Hallacy and
                  et al.},
  booktitle    = {ICML},
  year         = {2021},
}

@inproceedings{jiang2023cross,
  title        = {Cross-Modal Implicit Relation Reasoning and Aligning for Text-to-Image Person Retrieval},
  author       = {D. Jiang and
                  M. Ye},
  booktitle    = {CVPR},
  year         = {2023}
}

@article{yan2023clip,
  title        = {Clip-driven Fine-grained Text-Image Person Re-identification},
  author       = {S. Yan and
                  N. Dong and
                  L. Zhang and
                  et al.},
  journal      = {IEEE Trans. Image Process.},
  year         = {2023},
  publisher    = {IEEE}
}

@inproceedings{cao2024empirical,
  title        = {An Empirical Study of Clip for Text-based Person Search},
  author       = {M. Cao and
                  Y. Bai and
                  Z. Zeng and
                  et al.},
  booktitle    = {AAAI},
  year         = {2024}
}

@inproceedings{zhao2024unifying,
  title        = {Unifying Multi-Modal Uncertainty Modeling and Semantic Alignment for Text-to-Image Person Re-identification},
  author       = {Z. Zhao and
                  B. Liu and
                  Y. Lu and
                  et al.},
  booktitle    = {AAAI},
  year         = {2024}
}

@inproceedings{qin2024noisy,
  title        = {Noisy-Correspondence Learning for Text-to-Image Person Re-identification},
  author       = {Y. Qin and
                  Y. Chen and
                  D. Peng and
                  et al.},
  booktitle    = {CVPR},
  year         = {2024}
}

@inproceedings{zuo2024ufinebench,
  title        = {UFineBench: Towards Text-based Person Retrieval with Ultra-fine Granularity},
  author       = {J. Zuo and
                  H. Zhou and
                  Y. Nie and
                  et al.},
  booktitle    = {CVPR},
  year         = {2024}
}

@inproceedings{liao2020interpretable,
  title        = {Interpretable and Generalizable Person Re-identification with Query-Adaptive Convolution and Temporal Lifting},
  author       = {S. Liao and
                  L. Shao},
  booktitle    = {ECCV},
  year         = {2020},
  organization = {Springer}
}

@article{kim2026dico,
  title        = {DiCo: Disentangled Concept Representation for Text-to-Image Person Re-identification},
  author       = {G. Kim and
                  C. Eom},
  journal      = {Neurocomputing},
  year         = {2026},
}

@inproceedings{heritier2025exam,
  title        = {ExaM: Unsupervised Concept-based Representation Learning to Better Explain Models in Vision Tasks},
  author       = {M. Heritier and
                  D. Mekhazni and
                  C. Leblond-Menard and
                  et al.},
  booktitle    = {CVPR},
  year         = {2025}
}

@inproceedings{kim2018interpretability,
  title        = {Interpretability beyond Feature Attribution: Quantitative Testing with Concept Activation Vectors (TCAV)},
  author       = {B. Kim and
                  M. Wattenberg and
                  J. Gilmer and
                  et al.},
  booktitle    = {ICML},
  year         = {2018},
}

@inproceedings{alehdaghi2025beyond,
  title        = {Beyond Patches: Mining Interpretable Part-Prototypes for Explainable AI},
  author       = {M. Alehdaghi and
                  R. Bhattacharya and
                  P. Shamsolmoali and
                  et al.},
  booktitle    = {AAAI},
  year         = {2026}
}

@article{chen2019looks,
  title        = {This Looks Like That: Deep Learning for Interpretable Image Recognition},
  author       = {C. Chen and
                  O. Li and
                  D. Tao and
                  et al.},
  journal      = {Adv. Neural Inf. Process. Syst.},
  year         = {2019}
}

@inproceedings{nauta2023pip,
  title        = {Pip-net: Patch-based Intuitive Prototypes for Interpretable Image Classification},
  author       = {M. Nauta and
                  J. Schlotterer and
                  M. Van Keulen and
                  et al.},
  booktitle    = {CVPR},
  year         = {2023}
}

@inproceedings{rstpreid,
  title        = {DSSL: Deep Surroundings-person Separation Learning for Text-based Person Retrieval},
  author       = {A. Zhu and
                  Z. Wang and
                  Y. Li and
                  et al.},
  booktitle    = {MM},
  year         = {2021},
}

@article{yang2025qwen3,
  title        = {Qwen3 Technical Report},
  author       = {A. Yang and
                  A. Li and
                  B. Yang and
                  et al.},
  journal      = {CoRR},
  volume       = {arXiv:2505.09388},
  year         = {2025}
}

@inproceedings{yang2023towards,
  title        = {Towards Unified Text-based Person Retrieval: A Large-scale Multi-attribute and Language Search Benchmark},
  author       = {S. Yang and
                  Y. Zhou and
                  Z. Zheng and
                  et al.},
  booktitle    = {MM},
  year         = {2023}
}

@inproceedings{tan2024harnessing,
  title        = {Harnessing the Power of MLLMs for Transferable Text-to-Image Person ReID},
  author       = {W. Tan and
                  C. Ding and
                  J. Jiang and
                  et al.},
  booktitle    = {CVPR},
  year         = {2024}
}

@inproceedings{chefer2021generic,
  title        = {Generic Attention-model Explainability for Interpreting Bi-modal and Encoder-Decoder Transformers},
  author       = {H. Chefer and
                  S. Gur and
                  L. Wolf},
  booktitle    = {ICCV},
  year         = {2021}
}

@article{ergasti2024mars,
  title        = {MARS: Paying More Attention to Visual Attributes for Text-based Person Search},
  author       = {A. Ergasti and
                  T. Fontanini and
                  C. Ferrari and
                  et al.},
  journal      = {ACM Trans. Multimedia Comput. Commun. Appl.},
  year         = {2025},
}
}

\clearpage
\appendix

\numberwithin{equation}{section}
\numberwithin{figure}{section}
\numberwithin{table}{section}

\clearpage

\twocolumn[
\begin{center}
{\Large \bfseries Supplementary Material}

\vspace{1em}
{\large \method: Phrase-Region Grounding for Interpretable Text-to-Image \\Person Re-Identification}
\end{center}

\vspace{2em}
]

\maketitle

\noindent This supplementary material contains the following content:
\begin{enumerate}[label=\textbf{\Alph*.},itemsep=0pt, parsep=0pt]
    \item \textbf{Extended Related Work.}
    This section provides a broader discussion of text-to-image person re-identification (TI-ReID) in addition to related work presented in the main paper.

    \item \textbf{Extended Evaluation Protocol.}
    This section details the experimental setting, including implementation details, phrase generation, optimization strategy, and evaluation measures.

    \item \textbf{Extended Results and Analysis.}
    This section presents the full retrieval results on CUHK-PEDES, ICFG-PEDES, and RSTPReID, along with additional sensitivity analyses of perturbation strength and relevance threshold under the proposed masking framework.

    \item \textbf{Sample Phrase Annotations.}
    This section provides representative examples of the phrase annotations used to supervise the proposed patch-phrase interaction module (PPIM), along with a brief description of the annotation schema and the decomposition of free-form captions into compact, visually grounded phrase units.

    \item \textbf{Code.} The code is attached as a zip file along with this document.
\end{enumerate}

\section{Extended Related Work}
\textbf{Text-to-image (TI-ReID).} TI-ReID aims to retrieve a target person's image from a large gallery using a free-form natural language description. Early approaches~\cite{gao2021contextual, li2017person, zhang2018deep} utilized modality-specific visual and textual encoders. These models are trained by optimizing global matching objectives to construct a shared embedding space. While these methods established the foundational cross-modal retrieval pipeline, they often failed to fully exploit fine-grained appearance cues, which are essential for distinguishing visually similar pedestrians. To address this limitation, subsequent research focuses on part- and region-aware alignment. For instance, PMA~\cite{jing2020pose} employed pose keypoints to define body regions. ViTAA~\cite{wang2020vitaa} exploits human parsing to separate attribute-level features, and SSAN~\cite{ding2021semantically} modelled implicit part correspondences without explicit pose or parsing supervision. Although these methods demonstrated the importance of local correspondence modelling, many still rely on handcrafted body decomposition or auxiliary structural priors.

Different methods have been proposed to improve the architecture, shifting from modality-specific encoders to general vision-language pre-trained backbones. For instance, CLIP-based vision-language models~\cite{radford2021learning} have enabled methods such as IRRA~\cite{jiang2023cross}, CFine~\cite{yan2023clip}, and TBPS-CLIP~\cite{cao2024empirical} to dramatically improve the cross-modal retrieval by leveraging stronger image-text priors and more effective fine-grained matching strategies. Recently, MUM~\cite{zhao2024unifying} models multimodal uncertainty to better handle ambiguous correspondences. RDE~\cite{qin2024noisy} directly addresses noisy image-text pairings during training, and ReID-domain pre-training methods such as APTM~\cite{yang2023towards} and NAM~\cite{tan2024harnessing} leverage domain-adapted large-scale supervision or MLLM-assisted annotation to improve robustness and generalization. Nevertheless, most of these methods remain focused on offline alignment and offer limited interpretability regarding which phrase-level evidence contributes to retrieval outcomes.

Recent TI-ReID research highlights two complementary directions. ICL~\cite{qin2025human} introduces an MLLM-driven interactive framework that refines ambiguous queries through test-time human-centred dialogue and re-ranking. ICL demonstrates that external multimodal knowledge can enhance retrieval under dynamic user inputs. In contrast, DiCo~\cite{kim2026dico} operates entirely offline and learns hierarchical correspondence via modality-shared slots and concept-level factorization, facilitating alignment from global identity cues to finer semantic properties such as colour, texture, and shape. These improvements show that TI-ReID is progressing beyond basic global alignment toward either interaction-assisted retrieval or more structured fine-grained representation learning. However, ICL requires additional test-time interaction with an MLLM, while DiCo relies on slot-based decomposition and does not explicitly ground open-vocabulary textual phrases to localized visual regions. In contrast to these approaches, our method directly promotes phrase-region alignment within a standard CLIP-based TI-ReID framework, aiming to achieve both competitive retrieval performance and explicit, faithful part-level grounding.

\section{Extended Evaluation Protocol}

\noindent \textbf{Implementation details.} Our model follows the RDE~\cite{qin2024noisy} protocol for dataset splits, evaluation, and preprocessing. The visual and text encoders are a CLIP ViT-B/16~\cite{radford2021learning} and the patch embeddings $\{\bm{z}_k\}_{k=1}^{K}$ are extracted from the final transformer layer, excluding \texttt{[CLS]}. Phrases are generated offline via Qwen3~\cite{yang2025qwen3} under a structured JSON schema, yielding $P$ phrases per caption; each phrase is encoded through $E_T$. Furthermore, models for all datasets are trained for 100 epochs with a batch size of 128 on a single NVIDIA H100. We employed AdamW with a cosine learning rate schedule and a 5-epoch linear warm-up; the peak learning rate and all hyperparameters are selected via random search on a validation set. Inference, phrase-level explanation maps are obtained by reshaping raw phrase-to-patch similarity scores into a spatial grid, requiring no additional network components. Counterfactual masking experiments zero out the highest-scoring patches per phrase across all perturbation analyses.

%%%%%%%%%%%%%%%%%%%%%%%%%%%%%%%%%%
\noindent \textbf{Evaluation Measures. } We evaluate retrieval performance using standard TI-ReID metrics: Rank-1 accuracy and mAP are reported in the main paper, with Rank-5, Rank-10, and mINP provided in the supplementary material. Beyond retrieval accuracy, we introduce a counterfactual region-removal protocol to quantitatively assess explanation faithfulness.
%%%%%%%%%%%%%%%%%%%%%%%%%%%%%%%%

\section{Extended Results and Analysis}
Tab.\ref{tab_main_ext}, extends the retrieval results as they are presented in the paper. As shown in the table, across all metrics and datasets, our method, \method{}, achieves competitive performance while being interpretable and, in some cases, marginally exceeds the existing state of the art.  
\begin{table*}[!h]
\centering 
\setlength{\abovecaptionskip}{0.1cm}
\arrayrulecolor{black} 
\resizebox{\textwidth}{!}{
\setlength\tabcolsep{1.5pt}
\begin{tabular}{l|cc|c|ccccc|ccccc|ccccc}\hline\hline
\rowcolor{gray!25}   &    &    &   & \multicolumn{5}{c|}{{CUHK-PEDES}}       & \multicolumn{5}{c|}{{ICFG-PEDES}}       & \multicolumn{5}{c}{{RSTPReid}}                          \\
\rowcolor{gray!25} Methods &Image Enc.&Text Enc. & Interp. & Rank-1 & Rank-5 & Rank-10 & mAP & mINP
& Rank-1 & Rank-5 & Rank-10 & mAP & mINP
& Rank-1 & Rank-5 & Rank-10 & mAP & mINP \\\hline
IRRA$^\flat$~\cite{jiang2023cross} \textit{(cvpr'23)} & CLIP-ViT & CLIP-X. & $\times$ & 74.05 & 89.48 & 93.64 & 66.57 & - & 64.37 & 80.75 & 86.12 & 38.85 & - & 61.90 & 80.60 & 89.30 & 48.08 & - \\
APTM~\cite{yang2023towards} \textit{(mm'24)} & Swin-B & BERT & $\times$ & 76.53 & 90.04 & 94.15 & 66.91 & - & 68.51 & 82.99 & 87.56 & 41.22 & - & 67.50 & 85.70 & 91.45 & 52.56 & - \\
NAM$^\natural$~\cite{tan2024harnessing} \textit{(cvpr'24)} & CLIP-ViT & CLIP-X. & $\times$ & 77.47 & 90.84 & 94.67 & 69.43 & \textbf{54.08} & 66.76 & 82.02 & 87.17 & 41.45 & \textbf{9.53} & 67.15 & 86.55 & 91.90 & 52.00 & 28.46 \\
PLOT~\cite{park2024plot} \textit{(eccv'24)} & CLIP-ViT & CLIP-X. & $\checkmark$ & 75.28 & 90.42 & 94.12 & - & - & 65.76 & 81.39 & 86.73  & - & - & 61.80 & 82.85 & 89.45 & - & - \\
ICL~\cite{qin2025human} \textit{(cvpr'25)} & CLIP-ViT & CLIP-X. & $\times$ & \textbf{78.18} & 91.63 & 94.83 & 69.58 & 53.48 & 69.22 & 83.49 & 88.06 & 42.34 & 9.01 & 70.00 & \textbf{86.60} & 91.70 & \textbf{54.16} & \textbf{30.93} \\
MARS~\cite{ergasti2024mars} (\textit{tomm'25})    & Swin-B     & BERT  & $\times$   & 77.62 & 90.63 & 94.27 & \textbf{71.71} & -     & 67.60 & 81.47 & 85.79 & \textbf{44.93} & -    & 67.60 & 81.47 & 85.79 & 44.93 & - \\
DiCO~\cite{kim2026dico} \textit{(n-computing'26)} & CLIP-ViT & CLIP-X. & $\checkmark$ & 77.21 & \textbf{91.85} & \textbf{95.63} & - & - & 67.81 & 83.29 &  87.62 & - & - & 67.84 & 85.72 & \textbf{91.98} & - & - \\
\hline
\textbf{\method} & CLIP-ViT & CLIP-X. & $\checkmark$ & 78.28 & 91.27 & 94.57 & 69.26 & 52.87 & \textbf{69.92} & \textbf{83.83} & \textbf{88.14} & 42.03 & 8.79 & \textbf{71.05} & 85.50 & 90.80 & 52.76 & 28.00 \\
\hline\hline
\end{tabular} 
} 
\caption{Performance on the three coarse-grained benchmarks.} 
\label{tab_main_ext}
\end{table*}

\begin{figure*}[h]
    \centering

    \begin{subfigure}{\linewidth}
        \centering
        \begin{overpic}[width=1\linewidth,trim=0cm 0cm 0cm 1cm, clip]{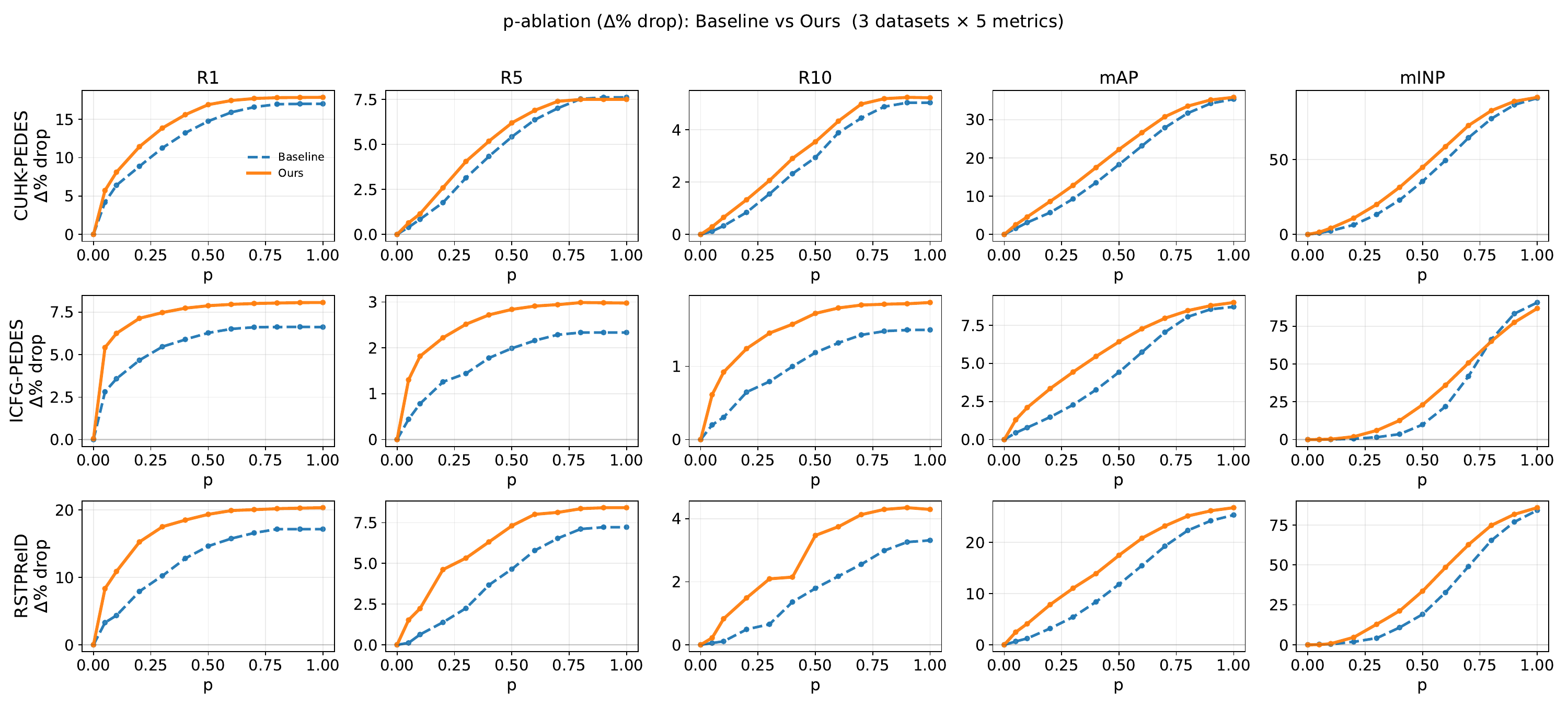}
            \put(50,90){\makebox(0,0){\colorbox{white}{}}}
        \end{overpic}
        \caption{}
        \label{fig:attention_vis_p}
    \end{subfigure}

    \begin{subfigure}{\linewidth}
        \centering
        \begin{overpic}[width=1\linewidth,trim=0cm 0cm 0cm 1cm, clip]{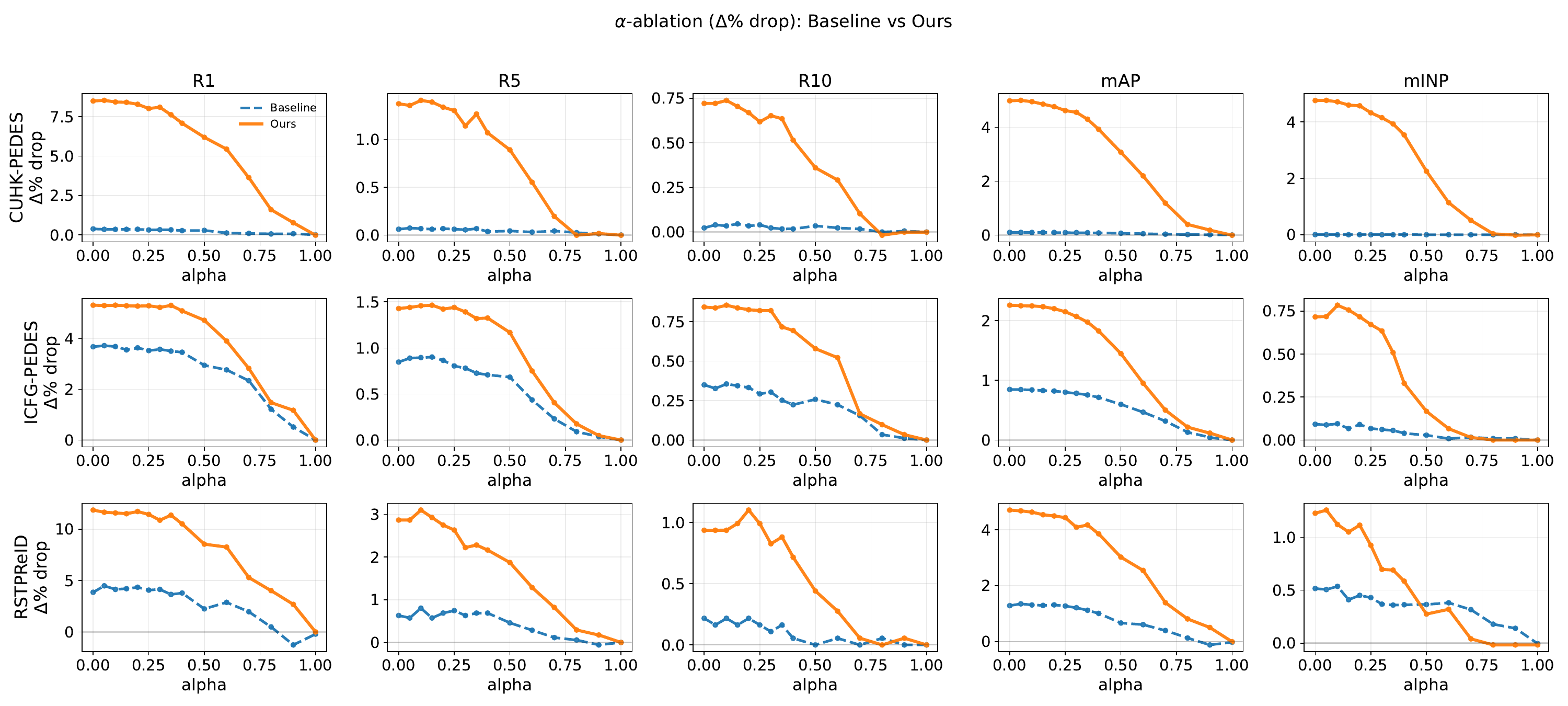}
            \put(50,90){\makebox(0,0){\colorbox{white}{}}}
        \end{overpic}
        \caption{
        }
        \label{fig:attention_vis_a}
    \end{subfigure}
    \caption{\textbf{Sensitivity analysis of relevance-based masking.}
    \textbf{(a)} Sensitivity to perturbation strength $p$, with fixed $\alpha =0.3$.
    Relative drops ($\Delta\%$) in all metrics when progressively masking larger fractions $p$ of the most influential phrase region.
    \textbf{(b)} Sensitivity to thresholded relevance masking $\alpha$.
    Relative metric drops when the relevance threshold $\alpha$ varies while the masking ratio is fixed ($p=0.1$).
    }
\end{figure*}

Fig.\ref{fig:attention_vis_p} shows how the retrieval performance drops as we increase the perturbation strength. Perturbation strength parameter, $p$, controls the fraction of the thresholded region (controlled via $\alpha$ parameter) that is to be perturbed (information removed, or disguised). As can be seen across all five metric and three datasets, our method, \method{} consistently achieves higher performance drops across varied perturbation strength. This shows that our method is more faithful. At a given $p$, the proposed interpretable units by our model are more reliable and accurate, because upon removing those top-ranked interpretable parts, leads to the model's retrieval performance degrading far more than if we remove the top-ranked interpretable part propsed by the baseline model. Similarly the thresholing parameter $\alpha$ controls the size of the proposed region (interpretable units). In a normalized feature map, $\alpha$ controls what part of the entire feature map is going to be proposed as the interpretable part. After $\alpha$ proposes the interpretable part, $p$ controls what part (fraction) of the interpretable part is being perturbed. For example, Fig.\ref{fig:attention_vis_p}, the plots use a fixed $\alpha = 0.3$, meaning that all the pixels corresponding to values higher than $0.3$ are being proposed as interpretable parts. And for Fig.\ref{fig:attention_vis_a}, the plots use a fixed $p = 0.1$, meaning that from the proposed thresholded region, $10\%$ of that region is being perturbed.

\section{Sample Phrase Annotations}\label{sec:supp_phrase_annotations}
This section provides some examples of the phrase annotations used to supervise the proposed patch-phrase interaction module (PPIM). As described in Sec.~3, each training sample is associated with an unordered set of short
appearance phrases $L_i=\{\ell_{i,1}, \ldots, \ell_{i,P}\}$ generated offline using a Qwen3-VL annotator under a constrained JSON schema. The goal is not to rewrite the full caption, but to decompose each image--text pair into compact,
visually grounded units that can be aligned with localized image patches during training. These annotations act only as lightweight auxiliary supervision; the original captions remain unchanged for the base global retrieval objective.

\noindent \textbf{Annotation protocol.} The schema is lightweight and open-vocabulary. Each annotation is
categorized into four coarse groups: \texttt{clothing},
\texttt{accessories}, \texttt{body\_appearance}, and \texttt{hair\_head}.
These groups correspond to semantically meaningful person parts and attributes
that can be localized in the image. Within each field, the annotator produces
non-empty lists of short lowercase phrases, typically 1-4 words, such as
\texttt{black jacket}, \texttt{green vest}, or \texttt{red sneakers}. This
restriction reduces lexical fragmentation while preserving the local semantic
detail required for patch-phrase supervision. When evidence is absent,
ambiguous, or outside the visible crop, the annotation falls back to a
controlled token from \texttt{none}, \texttt{unknown}, or
\texttt{not\_visible}.

The schema also decomposes the annotations into different parts. The \texttt{clothing} group decomposes the person into upper-body, lower-body, and footwear phrases. The \texttt{accessories.backpack} field is represented by a single categorical entry, whereas \texttt{accessories.others} retains noun-level object phrases such as \texttt{shopping bag}, \texttt{hat}, or \texttt{flag}. The \texttt{body\_appearance} group stores one compact phrase for gender, pose/motion, and height impression, and \texttt{hair\_head} combines a short
hair descriptor with optional head cues such as hat and glasses. Although the labels are serialized under coarse groups for readability and normalization, the training objective ultimately consumes the resulting unordered set of valid phrases rather than a fixed slot ordering.

\noindent\textbf{Annotation quality.}
Tab.~\ref{tab:supp_fault} quantifies the reliability of the generated phrase annotations across the three benchmarks. We define a phrase \emph{field} as a single schema slot (for example, \texttt{clothing.upper\_body} or \texttt{hair\_head}), and we count a field as \emph{faulty} when the annotator cannot ground it in visible evidence, namely when it is left unpopulated or filled with one of the controlled fallback tokens \texttt{none}, \texttt{unknown}, or \texttt{not\_visible}. The fault rate is then the fraction of faulty fields over all sample$\times$category field pairs in the training split. This rate is stable across datasets, ranging from $16.47\%$ on CUHK-PEDES to $17.51\%$ on RSTPReID, with ICFG-PEDES in between at $16.97\%$. The mild increase on RSTPReID is consistent with its shorter captions and smaller crops, which expose fewer groundable attributes per image. After discarding faulty fields, roughly $10$ valid phrases remain per sample (last column), providing a dense supervisory signal for the patch-phrase objective relative to the small fixed number of slots used by slot-attention baselines. Because faulty fields are masked at the loss level ($m_{i,p}{=}0$) and genuine concepts recur across the multiple images of each identity, this level of annotation noise does not harm training stability, as analysed in the robustness study of the main text (Sec.~5.3).

\begin{table}[h]
\centering
\setlength{\tabcolsep}{6pt}
\begin{tabular}{lccc}
\toprule
\textbf{Dataset} & \textbf{Train} & \textbf{Fault (\%)} & \textbf{Valid phr./samp.} \\
\midrule
CUHK-PEDES & 34{,}054 & 16.47 & 10.1 \\
ICFG-PEDES & 34{,}674 & 16.97 & 10.0 \\
RSTPReID   & 18{,}505 & 17.51 & \phantom{0}9.9 \\
\bottomrule
\end{tabular}
\caption{\textbf{Phrase annotation quality.} The fault rate is computed over phrase \emph{fields} (sample$\times$category pairs), counting fields populated with a controlled fallback token (\texttt{none}, \texttt{unknown}, \texttt{not\_visible}) or left unpopulated. Roughly $10$ valid phrases per sample remain as effective supervision.}
\label{tab:supp_fault}
\end{table}

\noindent\textbf{Reproducibility details.}
Phrase annotations are generated offline with an OpenAI-compatible vLLM interface. For each item, the generator reads the original dataset caption(s) and all caption-like auxiliary evidence available in the annotation record, including \texttt{mllm\_captions}, \texttt{qa\_captions}, \texttt{mllm\_dec\_captions}, and \texttt{diversity\_captions}. These strings are deduplicated and packed into a single \texttt{all\_text\_evidence} block. For the image-conditioned setting, the full pedestrian crop is provided as visual evidence and the prompt explicitly gives priority to image pixels over
noisy textual evidence. The model is \texttt{Qwen/Qwen3-VL-30B-A3B-Instruct}, decoded with
\texttt{temperature=0.0} and \texttt{max\_tokens=700}. This does not mean that the generation itself is deterministic, and this is exactly why we perform a prompt sensitivity analysis (Tab.4 of the main text) using different prompts and their configurations (detailed below) and show that our method performs well across diverse generation configurations.

\noindent\textbf{Prompt versions.}\label{sec:prompt_versions}
The generator defines three prompt variants. \textbf{P0} is the archived strict image-conditioned prompt: it requires valid JSON only, exact schema compliance, non-empty lowercase string lists, 1--4 word phrases, image pixels as primary evidence, text as secondary evidence, and controlled fallback tokens. \textbf{P1}, used for the image-conditioned annotations reported here, retains the P0 schema and phrase constraints but adds conflict resolution (\emph{follow the image when text conflicts}), discourages rigid template phrasing, allows natural wording variation when equally supported, and does not invent attributes or mixing fallback tokens with positive tags. \textbf{P1-T} is the caption-only control: no image is provided, text evidence becomes the primary evidence, and underspecified attributes must be assigned controlled fallback tokens. The released records store the exact system prompt, exact user prompt, packed evidence block, model name, decoding parameters, image-evidence metadata, and image hashes in \texttt{json\_labels\_meta}.

The same schema and user-template are used for CUHK-PEDES, ICFG-PEDES, and RSTPReID; only the values inserted into the placeholders differ by dataset and sample. The prompt variant is selected by the annotation run and is recorded in
the metadata.

\noindent\textbf{System prompt templates.}
{\footnotesize
\begin{Verbatim}[breaklines=true,breakanywhere=true]
P0 (strict image-conditioned): you are an expert pedestrian attribute annotator. output only valid json (no markdown, no extra text). the output must match the required schema exactly (no extra keys). never output empty lists; every field must be a non-empty list of lowercase strings. each list item must be 1-4 words; prefer 2-4 words whenever possible. use image pixels as primary evidence; use provided text evidences as secondary hints. if unclear, use exactly one of: none, unknown, not_visible.

P1 (image-conditioned, used in the main generated labels): you are an expert pedestrian attribute annotator. output only valid json with no markdown, explanation, or extra text. the output must match the required schema exactly and contain no extra keys. every field must be a
non-empty list of lowercase strings. each list item must be 1-4 words, preferably 2-4 words when possible. use image pixels as the primary evidence and treat provided text as secondary evidence that may be noisy. if text conflicts with the image, follow the image. avoid overly rigid template wording. when multiple phrasings are equally supported, choose a natural valid phrasing rather than repeating the same stock phrase across samples. allow wording variation across examples, but keep the meaning faithful to the evidence and do not invent details. if an attribute is unclear, missing, or not visible, use exactly one of: none, unknown, not_visible. never mix fallback tokens with positive tags in the same list.

P1-T (caption-only control): you are an expert pedestrian attribute annotator. output only valid json (no markdown, no extra text). the output must match the required schema exactly (no extra keys). never output empty lists; every field must be a non-empty list of lowercase strings. each list item must be 1-4 words; prefer 2-4 words whenever possible. no image will be provided; use provided text evidences as primary evidence. if unclear, use exactly one of: none, unknown, not_visible.
\end{Verbatim}
}

\noindent\textbf{User prompt template.}
{\footnotesize
\begin{Verbatim}[breaklines=true,breakanywhere=true]
{IMAGE_OR_TEXT_ONLY_PREAMBLE}

caption_1: {CAPTION_1}
caption_2: {CAPTION_2}

all_text_evidence (deduplicated):
{EVIDENCE_LINES}
{OPTIONAL_EXISTING_JSON_LABELS}

OUTPUT JSON SCHEMA:
{
  "schema": ["chuk_peds_parts_phrase_v1"],
  "clothing": {
    "upper_body": ["unknown"],
    "lower_body": ["unknown"],
    "footwear": ["unknown footwear"]
  },
  "accessories": {
    "backpack": ["unknown"],
    "others": ["unknown"]
  },
  "body_appearance": {
    "gender": ["unknown"],
    "pose_motion": ["unknown"],
    "height_impression": ["unknown"]
  },
  "hair_head": ["unknown"]
}

Hard requirements: output only valid json; every field is a non-empty lowercase string list; each phrase is 1-4 words; use only none, unknown, or not_visible for missing evidence; never mix fallback and positive tags.

Phrase rules: clothing fields use descriptor+noun phrases; footwear also includes a coarse tone when visible; hair_head includes hair and optional hat/glasses cues; accessories are noun-only; body_appearance uses one phrase each for gender, pose_motion, and height_impression.
}

Representative examples.
We report representative entries whose captions and part labels are visually consistent. Each example includes the dataset caption(s), final json_labels, and the compact metadata fields that reproduce the generation setting. Full prompts and packed evidence are stored in json_labels_meta in the released JSON files.
\end{Verbatim}
}
\medskip
\noindent\textbf{Example 1, CUHK-PEDES p8130\_s10938.jpg}

\noindent\textbf{\texttt{caption\_1}:} The man is wearing safety gear including a neon yellow vest, a blue button-up shirt with an official logo on the left arm, black pants, black shoes, a black and red hat with a logo on the front, and he is carrying a red wand for directing traffic.

\noindent\textbf{\texttt{caption\_2}:} the man is wearing an officer hat, a safety vest that is a bright yellow neon color. he has a light blue button up short sleeve shirt with a pair of black pants and black shoes. He is facing the camera and has a red object in his right hand and a black item in his left hand.

{\footnotesize
\begin{Verbatim}
{
  "schema": ["chuk_peds_parts_phrase_v1"],
  "clothing": {
    "upper_body": [
      "light blue shirt",
      "neon yellow vest"
    ],
    "lower_body": ["black pants"],
    "footwear": [
      "dark footwear"
    ]
  },
  "accessories": {
    "backpack": ["none"],
    "others": [
      "hat",
      "flag"
    ]
  },
  "body_appearance": {
    "gender": ["male"],
    "pose_motion": ["standing"],
    "height_impression": ["average height"]
  },
  "hair_head": [
    "short hair",
    "wearing hat",
    "no glasses"
  ]
}
\end{Verbatim}
}
\noindent\textbf{\texttt{json\_labels\_meta} or input:}
{\footnotesize
\begin{Verbatim}[breaklines=true,breakanywhere=true]
{
  "model": "Qwen/Qwen3-VL-30B-A3B-Instruct",
  "temperature": 0.0,
  "max_tokens": 700,
  "include_images": true,
  "image_evidence": [{
    "label": "full image",
    "sha256": "<stored in released metadata>"
  }]
}
\end{Verbatim}
}

\medskip
\noindent\textbf{Example 2, ICFG-PEDES 
\\(\texttt{0627\_010\_05\_0303afternoon\_1591\_0.jpg}).
}\\
\textbf{\texttt{caption\_1}:} A young age woman has black shoulder-length hair and her is wearing a long grey collar jacket with white rounded neck inside with black pants and black shoes.
\textbf{\texttt{caption\_2:}:} Not applicable; ICFG-PEDES provides one caption per image.

\smallskip
\noindent\textbf{\texttt{json\_labels}:}
{\footnotesize
\begin{Verbatim}[breaklines=true,breakanywhere=true]
{
  "schema": ["chuk_peds_parts_phrase_v1"],
  "clothing": {
    "upper_body": [
      "grey jacket",
      "white top"
    ],
    "lower_body": ["black pants"],
    "footwear": ["black shoes"]
  },
  "accessories": {
    "backpack": ["none"],
    "others": ["none"]
  },
  "body_appearance": {
    "gender": ["female"],
    "pose_motion": ["walking"],
    "height_impression": ["unknown"]
  },
  "hair_head": [
    "long dark hair",
    "no hat",
    "no glasses"
  ]
}
\end{Verbatim}
}
\noindent\textbf{\texttt{json\_labels\_meta} for input:}
{\footnotesize
\begin{Verbatim}[breaklines=true,breakanywhere=true]
{
  "model": "Qwen/Qwen3-VL-30B-A3B-Instruct",
  "temperature": 0.0,
  "max_tokens": 700,
  "include_images": true,
  "image_evidence": [{
    "label": "full image",
    "sha256": "<stored in released metadata>"
  }]
}
\end{Verbatim}
}

\medskip
\noindent\textbf{Example 3, RSTPReID (\texttt{0005\_c5\_0024.jpg}).}\\
\textbf{\texttt{caption\_1}:} The man is walking, smiling and looking around. He is wearing a black leather jacket, dark blue jeans and black shoes. He is wearing a pair of glasses. He has short black hair.\\
\textbf{\texttt{caption\_2}:} The man is a little fat. He wears a pair of glasses. His black jacket looks a little tight. He wears a pair of navy-blue jeans. His shoes are black.

\smallskip
\noindent\textbf{\texttt{json\_labels}:}
{\footnotesize
\begin{Verbatim}[breaklines=true,breakanywhere=true]
{
  "schema": ["chuk_peds_parts_phrase_v1"],
  "clothing": {
    "upper_body": ["black leather jacket"],
    "lower_body": ["dark blue jeans"],
    "footwear": [
      "black shoes",
      "dark footwear"
    ]
  },
  "accessories": {
    "backpack": ["none"],
    "others": ["glasses"]
  },
  "body_appearance": {
    "gender": ["male"],
    "pose_motion": ["walking"],
    "height_impression": ["unknown"]
  },
  "hair_head": [
    "short black hair",
    "wearing glasses",
    "no hat"
  ]
}
\end{Verbatim}
}
\noindent\textbf{\texttt{json\_labels\_meta} for input:}
{\footnotesize
\begin{Verbatim}[breaklines=true,breakanywhere=true]
{
  "model": "Qwen/Qwen3-VL-30B-A3B-Instruct",
  "temperature": 0.0,
  "max_tokens": 700,
  "include_images": true,
  "image_evidence": [{
    "label": "full image",
    "sha256": "<stored in released metadata>"
  }]
}
\end{Verbatim}
}

These examples illustrate the intended behavior of the annotation pipeline.
The labels are phrase-centric rather than sentence-like, making them suitable
for direct patch-level supervision. Multiple entries are retained within a
field only when they provide complementary, visually supported descriptions,
such as \texttt{light blue shirt} and \texttt{neon yellow vest} for the
upper-body region in Example~1. Ambiguous or unavailable information is
represented by controlled fallback tokens rather than invented attributes. The
earlier inconsistent example with \texttt{patterned pants} was caused by
auxiliary evidence in the packed prompt that mentioned a pants pattern while
the two displayed captions only said \texttt{black pants}; we therefore replace
it here with examples whose displayed captions, auxiliary evidence, and visual
labels are easier to audit. During training, labels marked as
\texttt{unknown}, \texttt{none}, or \texttt{not\_visible} are masked and do not
contribute to the phrase-level loss.

\end{document}